\documentclass[runningheads]{llncs}
\PassOptionsToPackage{svgnames}{xcolor}

\usepackage{eccv}
\usepackage{eccvabbrv}

\usepackage{graphicx}
\usepackage{booktabs}
\usepackage{caption}
\usepackage{enumitem}
\usepackage{xspace}
\usepackage{wrapfig}
\usepackage{url}
\usepackage{amsfonts}
\usepackage{nicefrac}
\usepackage{microtype}
\usepackage{xcolor}
\usepackage{amsmath}
\usepackage{tabularx}
\usepackage{amssymb}
\usepackage{multirow}
\usepackage{relsize}
\usepackage{array}
\usepackage{colortbl}
\usepackage{pifont}
\usepackage{hhline}
\usepackage{boldline}
\usepackage{float}
\usepackage{subcaption}
\usepackage[figuresright]{rotating}
\usepackage{balance}
\usepackage{blindtext}
\usepackage{algorithm}
\usepackage{algpseudocode}
\usepackage{arydshln}
\usepackage{listings}
\usepackage{adjustbox}
\usepackage[most]{tcolorbox}
\tcbuselibrary{listings,breakable}

\newcommand{\ourmethod}{{\textsc{VidPair-Halluc}}\xspace}

\usepackage{amsmath,amsfonts,bm}

\def\eqref#1{equation~\ref{#1}}

\def\1{\bm{1}}

\DeclareMathAlphabet{\mathsfit}{\encodingdefault}{\sfdefault}{m}{sl}
\SetMathAlphabet{\mathsfit}{bold}{\encodingdefault}{\sfdefault}{bx}{n}

\newcommand{\ind}{\mathbf{1}}

\AtBeginDocument{%
  \setlength{\textfloatsep}{10pt plus 2pt minus 2pt}%
  \setlength{\floatsep}{9pt plus 2pt minus 2pt}%
  \setlength{\intextsep}{9pt plus 2pt minus 2pt}%
  \setlength{\abovecaptionskip}{5pt}%
  \setlength{\belowcaptionskip}{3pt}%
  \setlength{\abovedisplayskip}{5pt plus 1pt minus 1pt}%
  \setlength{\belowdisplayskip}{5pt plus 1pt minus 1pt}%
}

\usepackage[hidelinks]{hyperref}
\usepackage{orcidlink}

\begin{document}

\title{No Place to Hide: Benchmarking Video Hallucination with Background-Controlled Pairs}
\titlerunning{No Place to Hide}

\author{%
\mbox{Haojian Huang}\inst{1,2}\quad
\mbox{Harold Haodong Chen}\inst{1,2}\quad
\mbox{Meng Luo}\inst{3}\\
\mbox{Junjia Du}\inst{4}\quad
\mbox{Shanqing Xu}\inst{5}\quad
\mbox{Ziheng Chen}\inst{6}\\
\mbox{Yanxiang Huang}\inst{7}\quad
\mbox{Yinchuan Li}\inst{2}\quad
\mbox{Ying-Cong Chen}\inst{1,2}\thanks{Corresponding author.}}

\authorrunning{H.~Huang et al.}

\institute{%
\scriptsize
\begin{tabular}{@{}c@{}}
\textsuperscript{1}\,The Hong Kong University of Science and Technology (Guangzhou), Guangzhou, China \\
\textsuperscript{2}\,Knowin AI, China \quad
\textsuperscript{7}\,The Hong Kong Polytechnic University, Hong Kong SAR, China \\
\textsuperscript{3}\,National University of Singapore, Singapore \quad
\textsuperscript{4}\,Nanyang Technological University, Singapore \\
\textsuperscript{5}\,Huazhong University of Science and Technology, Wuhan, China \\
\textsuperscript{6}\,University of Illinois Urbana-Champaign, Urbana, IL, United States
\end{tabular}}

\maketitle

\begin{abstract}
We introduce \ourmethod, a new benchmark for evaluating video hallucination in large video models (LVMs) under rigorous and controlled conditions. Unlike previous benchmarks that primarily rely on text-based perturbations or adversarial questions while neglecting the consistency of visual backgrounds, \ourmethod features video pairs with highly similar backgrounds but distinctly different foreground semantics, enabling precise attribution of model errors to genuine hallucination rather than background variation. The benchmark is constructed through \textsc{PairFlow}, a pipeline that leverages recent advances in text-to-image and video generation to systematically compose stories, generate coherent video clips, and assemble them into adversarial pairs. Covering both spatial and temporal reasoning across ten semantic aspects, VidPair-Halluc comprises $1$K high-quality adversarial video pairs and $11$K spatio-temporal QA pairs with control over background and foreground variations. Evaluations on mainstream LVMs show persistent difficulty with robust fine-grained video understanding in adversarial settings, and code and data are available at the \href{https://jethrojames.github.io/VidPair-Halluc/}{\textcolor{magenta}{project page}}.
\end{abstract}

\section{Introduction}
\label{sec:intro}
\begin{figure}[ht]
    \centering
    \includegraphics[width=0.95\linewidth]{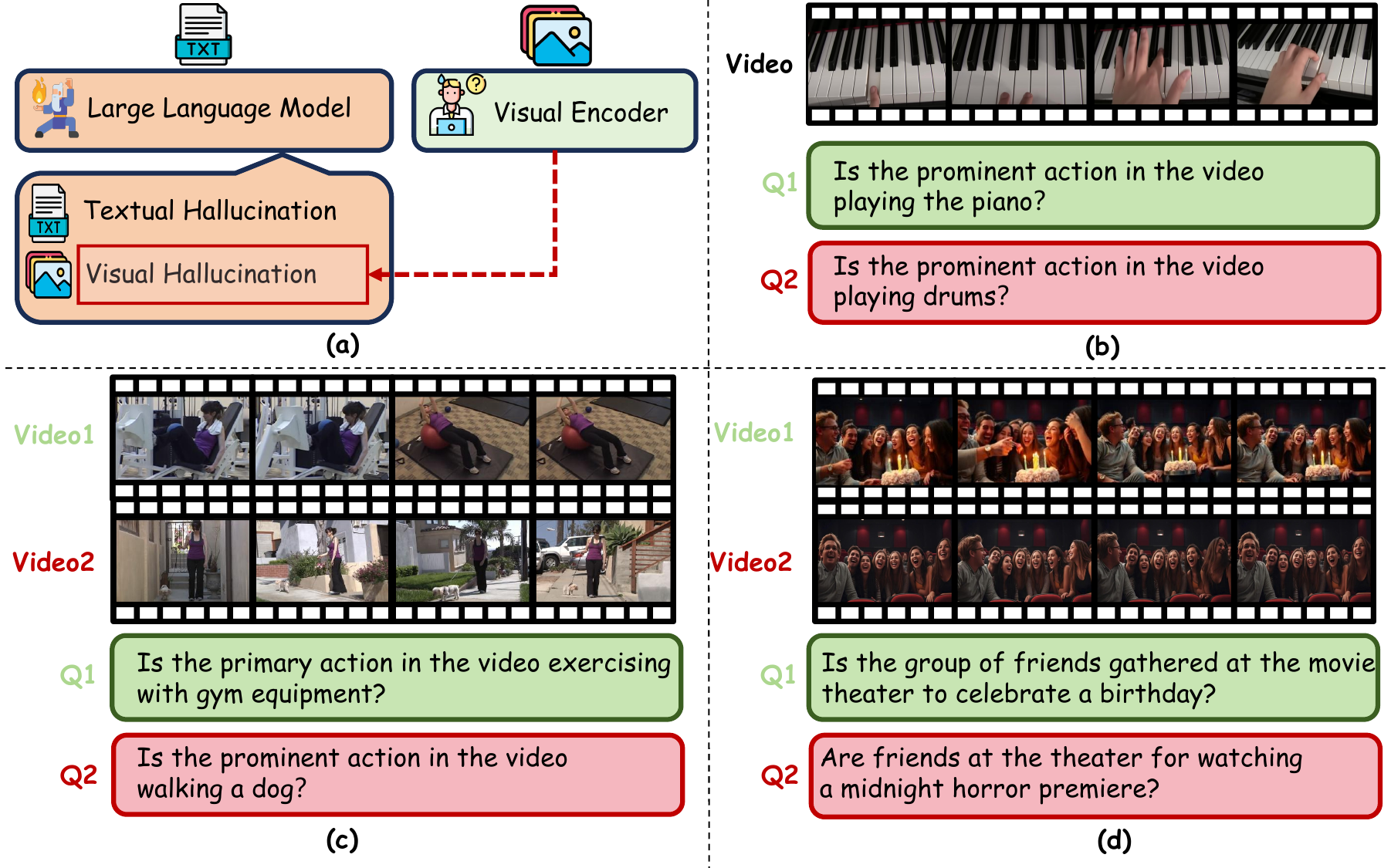}
    \caption{(\textbf{\textit{a}}) Hallucinations in LVMs, caused by the LLM or visual encoder, are categorized based on adversarial sources into textual or visual, where visual hallucination poses a greater challenge. (\textbf{\textit{b}}) Adversarial question benchmarks (\textit{e.g.}, VideoHallucer~\cite{wang2024videohallucer}) mainly induce textual hallucinations by perturbing the LLM. (\textbf{\textit{c}}) \textsc{VidHalluc}~\cite{li2024vidhalluc} introduces video pairs that are visually dissimilar but share similar overall semantics, challenging multimodal understanding. (\textbf{\textit{d}}) Our benchmark constructs video-text pairs with highly similar background but distinctly different foreground semantics, enabling targeted evaluation of hallucinations.}
    \label{fig:comp}
\end{figure}
Progress in video understanding is driven by large-scale datasets~\cite{fu2024video,li2024mvbench,wang2024videohallucer,zhou2024humanvbench,Bain21,xiao2021next,wang2019vatex} and advanced  large video models (LVMs)~\cite{lin2023video,bai2023qwen,cheng2024videollama,li2023videochat,maaz2023video,chen2024sharegpt4video}. A key challenge is video hallucination, where models generate content inconsistent with visual evidence, often due to a capability gap between video encoders and large language models (LLMs)~\cite{huang2025vistadpo,yuan2024videorefer}. While LLMs undergo extensive pre-training, weaker video encoders lead to overconfident yet inaccurate outputs~\cite{liang2024survey,cui2024robustness}. To address video hallucination, benchmarks have evolved progressively. Initial works~\cite{Vript,zhang2024eventhallusion,gao2025exploring} focus on single videos, using spatio-temporal questions to assess basic understanding. Building on this, subsequent studies introduce adversarial textual contexts (see Figure~\textcolor{magenta}{\ref{fig:comp}} (\textit{b})), such as misleading questions or distractor answers within single videos~\cite{wang2024videohallucer}, to more comprehensively evaluate model robustness to high-level semantic perturbations. Nevertheless, since most LVMs inherently depend on instruction tuning, their predisposition to strictly follow user directives suggests that hallucinations observed in these benchmarks may be confounded by human-induced bias, thereby undermining the objectivity of such evaluations. 
Recent benchmarks employ paired videos~\cite{guan2024hallusionbench,li2024vidhalluc} to assess fine-grained semantic alignment, yet they mainly rely on text-driven disturbances and CLIP/DINO-based selection without explicitly controlling background consistency. Consequently, hallucination behavior may be confounded by background variation rather than purely reflecting foreground semantic misinterpretations, where the background is query-irrelevant and the foreground query-relevant visual context. While background variation is intrinsic to real-world hallucinations, controlling for similarity serves as a controlled setting that enables more precise attribution to foreground semantics. Such isolation is crucial for disentangling sources of video hallucination, enabling a more systematic understanding of model robustness.

To address these limitations, we propose a hallucination benchmark with background-consistent yet foreground-divergent video pairs, enabling clearer attribution of model errors to hallucination rather than background shifts or semantic misguidance. However, constructing such adversarial pairs in real videos is labor-intensive, as fixed storylines and spatio-temporal dependencies require extensive human intervention. In contrast, image-based VQA has leveraged advances in image inpainting and automated pipelines~\cite{zhuang2024task,zhang2024countercurate,wu2024autohallusion,wu2025symmetrical,liu2024finecops, google2024nano_banana}, achieving scalable adversarial sample generation. Inspired by this, we look to recent progress in video generation and editing to address scalability in the video domain. Yet, existing approaches remain limited:
(\textbf{\textit{i}}) Direct video generation or frame replacement often disrupts background or entity consistency.
(\textbf{\textit{ii}}) Video editing lacks stability and fidelity in complex scenes.
(\textbf{\textit{iii}}) Generating video pairs from edited key frames and descriptions faces challenges from scarce frame-level captions and high annotation costs.

These challenges motivate us to explore whether advanced text-to-image and video generation models can streamline data collection and construct adversarial video pairs with minimal human intervention. To this end, we propose \textbf{\textsc{PairFlow}}, a pipeline of three stages:
\ding{182} \textbf{\textit{Story Composition}:} Automatically generate paired stories with controlled dependencies.
\ding{183} \textbf{\textit{Video Clip Generation}:} Synthesize coherent clips via advanced text-to-image (T2I) and video generation models.
\ding{184} \textbf{\textit{Video Assembly}:} Concatenate clips into adversarial video pairs for robust LVM evaluation.
Based on PairFlow, we construct video pairs with similar backgrounds but distinct foregrounds, and introduce \textbf{\ourmethod} (Figure~\textcolor{magenta}{\ref{fig:comp}} (\textit{d})) for rigorous hallucination benchmarking.
Our contributions are as follows:
\begin{itemize}
    \item We propose \textsc{PairFlow}, a pipeline that constructs background-consistent, foreground-divergent adversarial video pairs for fine-grained video hallucination evaluation.
    \item We introduce \ourmethod, a new benchmark covering both spatial and temporal reasoning from $10$ perspectives, featuring $1$K high-quality adversarial video pairs and $11$K spatio-temporal QA pairs under rigorous background and foreground control.
    \item We benchmark mainstream LVMs on \ourmethod, providing in-depth analysis of their strengths and limitations. We further assess state-of-the-art hallucination mitigation methods, revealing current approaches remain insufficient for robust fine-grained video understanding.
\end{itemize}


\section{Related Work}
\label{sec:related_work}

\subsection{Video Hallucination Benchmarks}
Video hallucination arises when an LVM produces confident responses that are not supported by the visual evidence. Early benchmarks mainly evaluate this behavior within a single video. Vript-HAL~\cite{Vript} and EventHallusion~\cite{zhang2024eventhallusion} ask spatio-temporal questions about object relations, actions, and event sequences, exposing failures in basic video grounding and event-level reasoning. Later work introduces stronger textual adversaries. VideoHallucer~\cite{wang2024videohallucer}, for example, builds paired binary questions in which one question is factual and the other is hallucinated, making it possible to diagnose whether a model follows misleading language priors rather than the video content.

Recent benchmarks further move from single-video evaluation to paired visual inputs. HallusionBench~\cite{guan2024hallusionbench} disentangles visual evidence from language priors through carefully constructed image-question pairs, while \textsc{VidHalluc}~\cite{li2024vidhalluc} assembles thousands of video pairs to probe temporal hallucination. These paired settings are closer to the counterfactual nature of hallucination diagnosis, but they often rely on naturally different videos or feature-based retrieval. As a result, the visual background, camera motion, and foreground semantics may all vary at once. This makes it difficult to attribute a model error to the queried foreground fact alone. In contrast, \ourmethod deliberately fixes the background as much as possible while changing the query-relevant foreground semantics, yielding a controlled setting that complements real-video benchmarks.

\subsection{Adversarial Video Pairs for Alignment}
Preference and contrastive alignment have become important tools for mitigating hallucination in MLLMs and LVMs. Textual preference optimization methods such as DPO~\cite{rafailov2024direct} and multimodal variants~\cite{liu2024mia,xie2024v,zhou2024aligning,wang2024mdpo} align model outputs with preferred responses, and video-oriented methods such as Hound-DPO~\cite{zhang2024direct}, PAMI-VDPO~\cite{ding2025pami}, and VistaDPO~\cite{huang2025vistadpo} demonstrate that visual preference pairs can improve video-language grounding. However, high-quality video preference pairs are expensive: real videos contain fixed storylines, complex temporal dependencies, and many non-target changes that require careful manual annotation.

Generative models provide a scalable alternative. Recent video generation and image/video editing systems~\cite{kong2024hunyuanvideo,wan2025,magi1,zhuang2024task,zhu2024instantswap,Soni2025LOCATEditGL,Chen2025InstructCLIPII,Shi2024SeedEditAI} make it increasingly practical to create fine-grained counterfactual samples with controlled local changes. Related work also spans multi-shot generation, multimodal affective understanding, cross-modal reliability, repository-dependency evaluation, temporal grounding, video repair, spatial reasoning, and embodied VLM robustness~\cite{zheng2024videogen,hu2024recent,huang2025trusted,huang2024crest,du2025dependeval,huang2025demr,li2025scagiqa,zhao2026star,huang2026affordance,huang2026find,zhao2025evoempirbench,zhang2026worldlines,wu2026robostressbench}. This opens a path toward reusable hard negatives for curation, calibration, reranking, and contrastive or preference training. \textsc{PairFlow} follows this direction, but uses synthesis primarily for evaluation: by controlling the non-target background and altering only the queried value or order, it isolates the hallucination source more cleanly than mined real-video pairs.

\begin{figure}[!t]
    \centering
    \includegraphics[width=\linewidth]{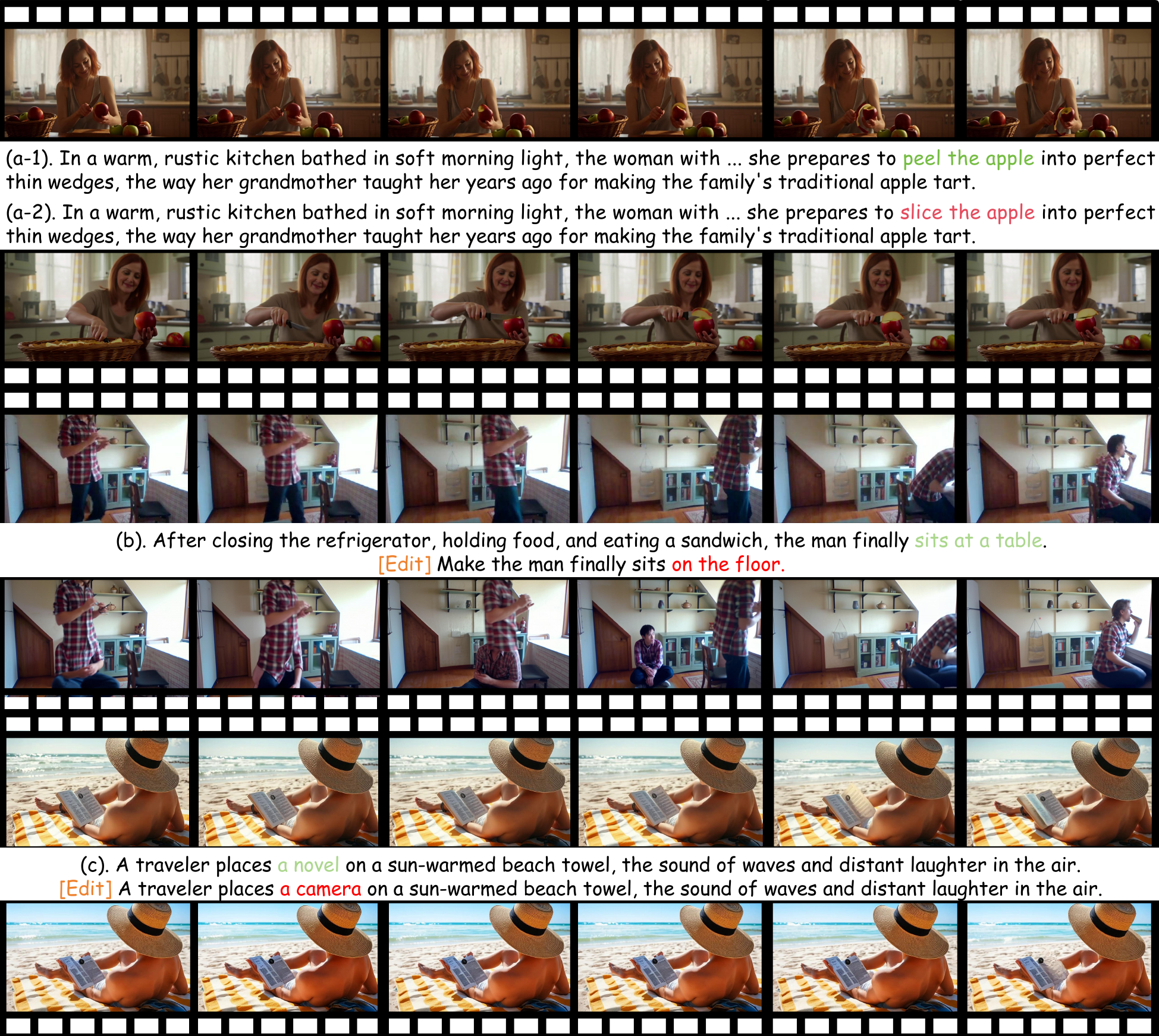}
    \caption{\textbf{Challenges in Generating High-Quality Adversarial Video Pairs}. (\textit{\textbf{a}}) Cutting-edge video generation models (\emph{i.e.} HunyuanVideo \cite{kong2024hunyuanvideo}) create pairs with significant semantic differences but fails to maintain consistent visual backgrounds. (\textit{\textbf{b}}) Advanced image editing methods (\emph{i.e.} SeedEdit \cite{Shi2024SeedEditAI}), applied frame-by-frame, lead to noticeable inconsistencies. (\textit{\textbf{c}}) Advanced video-editing techniques (\emph{i.e.} TokenFlow \cite{tokenflow2023}) result in pairs that lack faithfulness, stability and robustness.}
    \label{fig:comp_pair}
\end{figure}
\begin{figure*}[ht]
    \centering
    \includegraphics[width=0.96\linewidth]{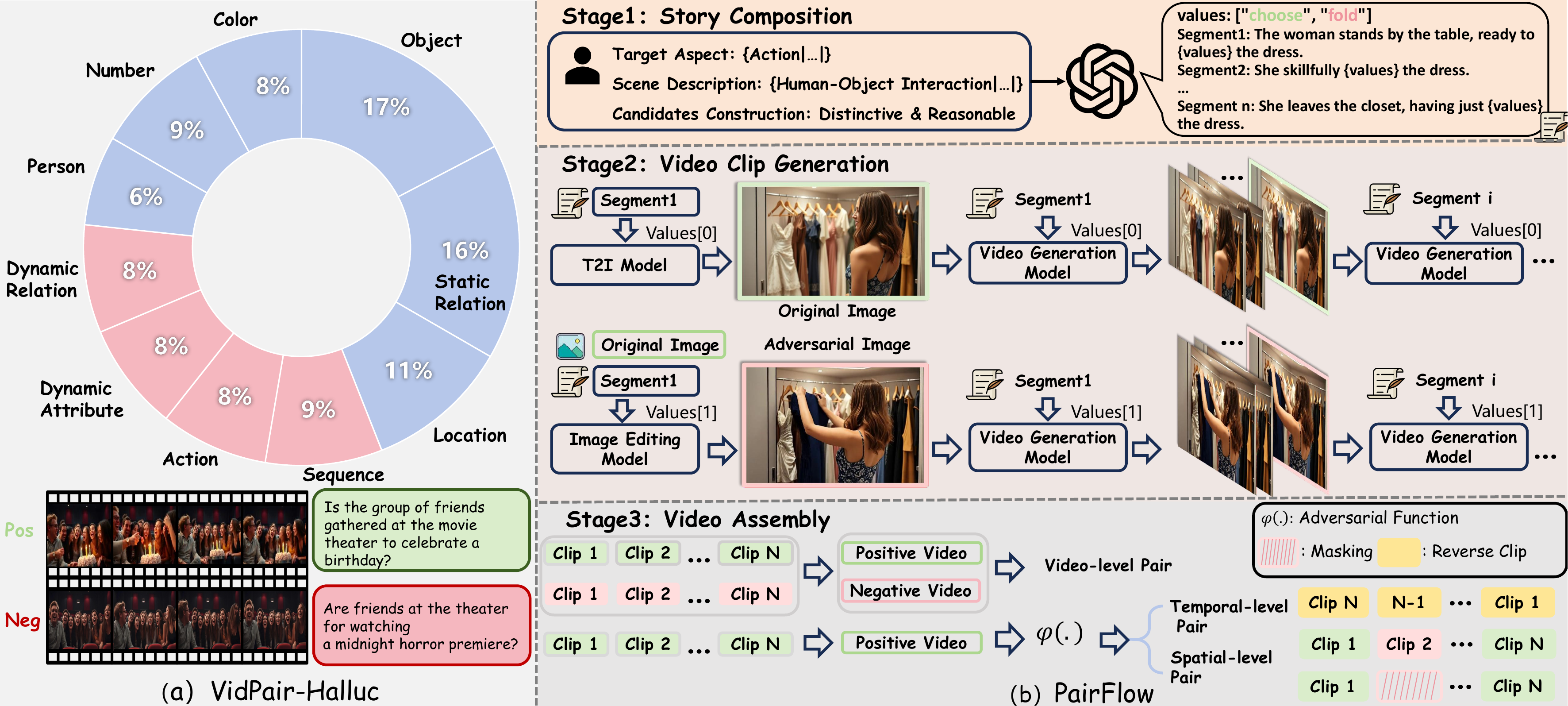}
    \caption{(\textbf{\textit{a}}) \ourmethod evaluates hallucinations from temporal ($33$\%) and spatial ($67$\%) aspects.
(\textbf{\textit{b}}) \textsc{PairFlow} generates video-text scenarios, synthesizes clips with controlled semantic variations, and assembles positive/negative pairs at multiple levels for robust hallucination benchmarking.}
    \label{fig:flowchart}
\end{figure*}
\section{\ourmethod: An Adversarial Benchmark for Video Hallucination}
\label{sec:method}
Existing benchmarks typically assess video hallucination either by posing misleading questions on a single video~\cite{wang2024videohallucer,zhang2024eventhallusion,choong2024vidhal} or by pairing videos with low visual similarity but distinct semantics~\cite{li2024vidhalluc,chen2024multi,guan2024hallusionbench}. A more demanding and challenging setting involves adversarial pairs that share highly similar backgrounds yet differ in entity semantics, where subtle distinctions are easily missed by LVMs~\cite{guan2024hallusionbench,Wu2025SymmetricalVC,chen2024multi}. However, constructing such pairs from real videos is costly and impractical, limiting current benchmarks' ability to probe hallucinations under these nuanced conditions. To close this gap, we build on PairFlow and introduce \ourmethod, which systematically evaluates video hallucination using hierarchical levels of adversarial video pairs.

\subsection{\textsc{PairFlow}: An Adversarial Video Pairs Data Pipeline}

\subsubsection{Motivations.} 
The fixed storylines and spatio-temporal dependencies in real videos demand extensive human effort, limiting the scalability of adversarial video pair generation with previous methods. In contrast, image-based VQA tasks have benefited from advances in image editing technologies~\cite{zhuang2024task,zhang2024countercurate}, enabling a shift from hand-crafted corner cases to automated pipelines that efficiently scale up adversarial samples generation~\cite{wu2024autohallusion,wu2025symmetrical,liu2024finecops}. Motivated by this progress, we turn to recent advances in video generation and editing as a promising path to overcome the scalability bottleneck in the video domain.

Fortunately, advances in video generative technologies indeed offer promising solutions to these challenges. Several intuitive candidate approaches can be envisioned: (\textit{\textbf{i}}) directly generating video pairs (\emph{e.g.}~\cite{bai2025impossible}) or editing frame by frame; (\textit{\textbf{ii}}) applying video editing (\emph{e.g.}~\cite{ma2025magicstick,zhou2023propainter}); (\textit{\textbf{iii}}) generating video pairs conditioned on edited key frames and descriptions. But these approaches still face several limitations:  
\ding{182} Simple video-level generation (\emph{i.e.} Figure~\textcolor{magenta}{\ref{fig:comp_pair}} (\textit{a})) or frame-level replacement methods (\emph{i.e.} Figure~\textcolor{magenta}{\ref{fig:comp_pair}} (\textit{b})) often fail to maintain entity or background consistency~\cite{tokenflow2023}, causing unnatural transitions and less plausible adversarial samples.
\ding{183} Current video editing techniques, though promising, still lack the stability and fidelity needed for reliable adversarial editing in complex scenes (\emph{i.e.} Figure~\textcolor{magenta}{\ref{fig:comp_pair}} (\textit{c})).
\ding{184} Editing-and-generation strategies are promising, but the lack of frame-level captions and varying frame resolutions in real videos hinder target selection, raising annotation costs and limiting scalability.

Given these observations, we are motivated to explore a more scalable and controllable paradigm. A key question arises: \textit{can advanced text-to-image and video generation models streamline data collection and efficiently construct adversarial video pairs with minimal human intervention?} To address this, we propose \textbf{\textsc{PairFlow}}, a novel framework to answer this by reducing manual annotation, while also holding the potential to enhance video-language alignment and video hallucination benchmarking. The details of data construction with \textsc{PairFlow} follows.
\begin{figure*}[t]
\centering
\includegraphics[width=0.96\linewidth]{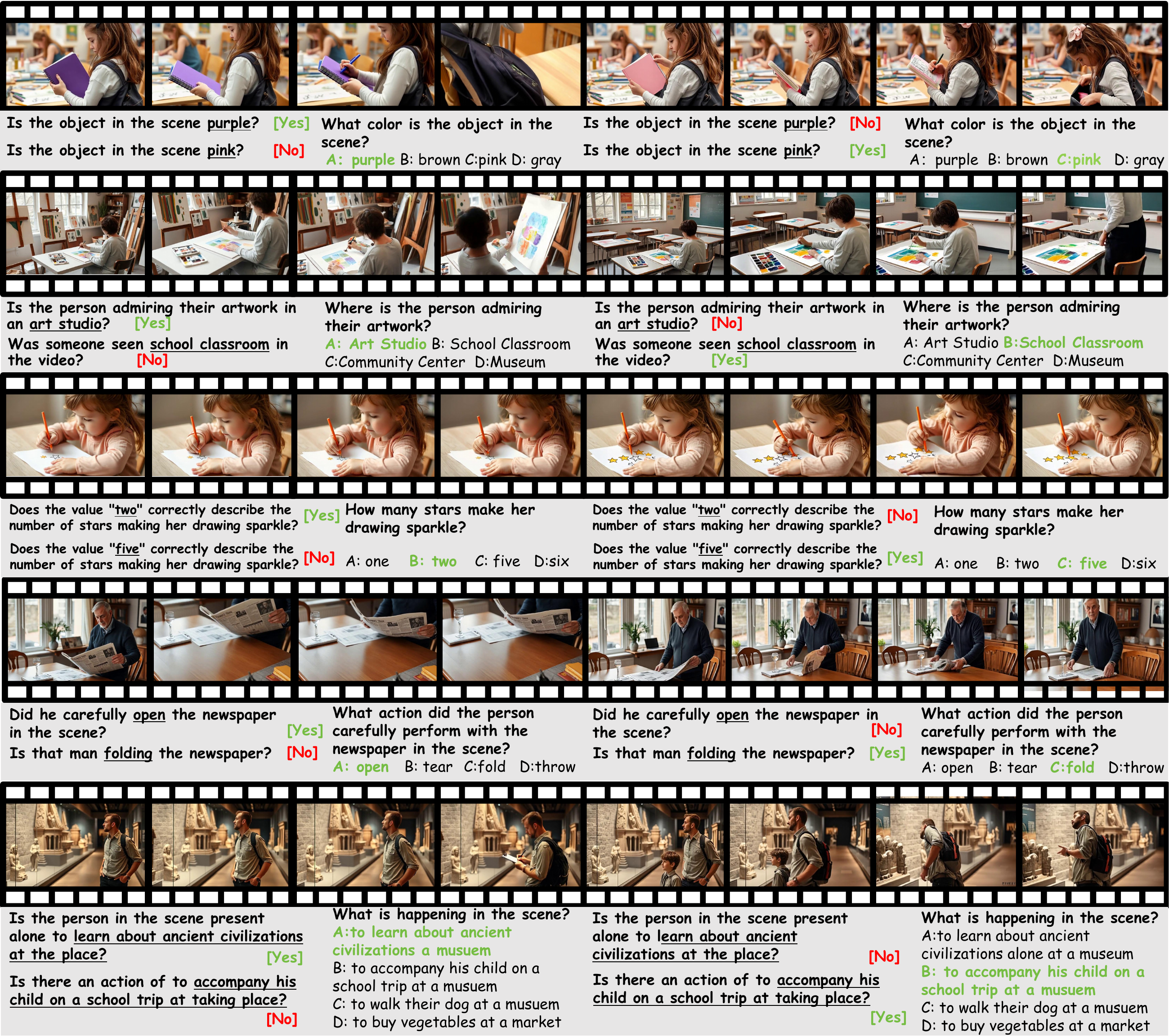}
\caption{Examples from \ourmethod. Each row shows a positive-negative video-level pair, with four frames per video (left: positive, right: negative). The first three rows illustrate spatial reasoning, while the last two focus on temporal reasoning. Binary QA and MCQ are provided for each pair, highlighting contrasting answers under highly similar visual contexts.}
\label{fig:video_level}
\end{figure*}

\subsubsection{Data Construction with \textsc{PairFlow}.}
PairFlow consists of three stages: story composition, video clip generation, and video assembly. Initially, the process begins with the composition of narratives, where specific elements such as actions and human-object interactions are emphasized. This stage involves constructing story segments using placeholders, allowing for the creation of diverse variations. For instance, a narrative might depict a woman preparing to perform an action like ``choose'' or ``fold'' a dress, represented as \textbf{\textrm{\{Values\}}}. Each candidate value is required to be logically consistent with the storyline and significantly different from one another.

In the video clip generation stage, an advanced pretrained text-to-image (T2I) model is employed to produce original images, which serve as the foundation for creating adversarial samples. These images undergo enhancement through sophisticated image editing techniques, ensuring that while the background remains consistent, the foreground exhibits significant distinction for high-quality adversarial samples. A noteworthy aspect of this process is the sequential generation of video segments, denoted as \( V_i \). Each segment \( V_i \) is generated using the last frame \( V_{i-1}^{L} \) of the preceding clip \( V_{i-1} \) in conjunction with the narrative script corresponding to segment \( V_i \). This approach is mathematically denoted as:
\begin{equation}
    V_i = f(V_{i-1}^{L}, S_i), \quad i \geq 2,
\end{equation}
where \( f(\cdot) \) denotes a video generation model which integrates the last frame \( V_{i-1}^{L} \) and the story script \( S_i \) to produce the subsequent video segment \( V_i \), ensuring each clip has a length of \( L \).

Finally, video assembly seamlessly integrates these clips into coherent sequences by crafting both positive and negative video versions through meticulous temporal and spatial pairings. The adversarial function $ \boldsymbol{\varphi(.)} $ is pivotal in this process, employing multiple operations to generate adversarial samples at both temporal and spatial levels. Temporally, each clip is individually reversed and then rearranged in reverse order \([V_N, V_{N-1}, \ldots, V_1]\), resulting in a completely inverted video sequence. Spatially, the process involves randomly replacing segments within video-level pairs and applying masking operations to the modified adversarial videos. Such adversarial samples pose significant challenges to the model's ability to comprehend specific foreground semantics across different clips.
\begingroup
\setlength{\tabcolsep}{1pt}
\renewcommand{\arraystretch}{1}
\begin{table}[t]
\centering
\caption{Comparison of existing video hallucination benchmarks.}
\scriptsize
\resizebox{\columnwidth}{!}{%
\begin{tabular}{lccccccc}
\toprule
\multirow{2}{*}{\textbf{Benchmark}} & \textbf{Number of} & \textbf{Binary} & \textbf{Multi-Choice} & \textbf{Open-Ended} & \textbf{Visual} & \textbf{Control} & \multirow{2}{*}{\textbf{Adversarial}} \\  
& \textbf{Question/Videos} & \textbf{QA} &  \textbf{QA} & \textbf{QA} & \textbf{Similarity} & \textbf{Pairs} &  \\
\midrule
HallusionBench~\cite{guan2024hallusionbench} & $1,129 / 346$ & \textcolor{ForestGreen}{\ding{51}} & \textcolor{ForestGreen}{\ding{51}} & \textcolor{FireBrick}{\ding{55}} & \textcolor{FireBrick}{\ding{55}} & \textcolor{ForestGreen}{\ding{51}} & \textcolor{ForestGreen}{\ding{51}} \\
VideoHallucer~\cite{wang2024videohallucer}  & $1,800 / 948$ & \textcolor{ForestGreen}{\ding{51}} & \textcolor{FireBrick}{\ding{55}}      & \textcolor{FireBrick}{\ding{55}} & \textcolor{FireBrick}{\ding{55}} & \textcolor{FireBrick}{\ding{55}}      & \textcolor{ForestGreen}{\ding{51}} \\
Vript-HAL~\cite{Vript}      & $122 / 122$   & \textcolor{FireBrick}{\ding{55}}     & \textcolor{ForestGreen}{\ding{51}} & \textcolor{FireBrick}{\ding{55}} & \textcolor{ForestGreen}{\ding{51}} & \textcolor{FireBrick}{\ding{55}}      & \textcolor{FireBrick}{\ding{55}} \\
EventHallusion~\cite{zhang2024eventhallusion} & -- $/ 400$   & \textcolor{ForestGreen}{\ding{51}} & \textcolor{FireBrick}{\ding{55}}      & \textcolor{ForestGreen}{\ding{51}} & \textcolor{FireBrick}{\ding{55}} & \textcolor{FireBrick}{\ding{55}} & \textcolor{FireBrick}{\ding{55}} \\
\textsc{VidHalluc}~\cite{li2024vidhalluc}        & $9,295 / 5,002$ & \textcolor{ForestGreen}{\ding{51}} & \textcolor{ForestGreen}{\ding{51}} & \textcolor{ForestGreen}{\ding{51}} & \textcolor{FireBrick}{\ding{55}} & \textcolor{ForestGreen}{\ding{51}} & \textcolor{ForestGreen}{\ding{51}} \\
\midrule
\rowcolor{cyan!10}
\textbf{\ourmethod (Ours)} & $11,523 / 2000$ & \textcolor{ForestGreen}{\ding{51}} & \textcolor{ForestGreen}{\ding{51}} & \textcolor{ForestGreen}{\ding{51}} & \textcolor{ForestGreen}{\ding{51}} & \textcolor{ForestGreen}{\ding{51}} & \textcolor{ForestGreen}{\ding{51}} \\
\bottomrule
\end{tabular}%
}

\label{tab:comp}
\end{table}
\endgroup

\begin{wraptable}{r}{0.5\textwidth}
\centering
\caption{Background Similarity Metrics.}
\resizebox{0.5\textwidth}{!}{%
\begin{tabular}{lccc}
\toprule
\textbf{Metric} & \textbf{Random} & \textbf{Non-Scene} & \textbf{Scene (Masked)} \\
\midrule
DINOv2 $\uparrow$ & 0.21$\pm$0.08 & 0.94$\pm$0.02 & 0.81$\pm$0.05 \\
LPIPS $\downarrow$ & 0.68$\pm$0.12 & 0.15$\pm$0.03 & 0.15$\pm$0.02 \\
SSIM $\uparrow$ & 0.24$\pm$0.06 & 0.76$\pm$0.05 & 0.90$\pm$0.04 \\
\bottomrule
\end{tabular}%
}
\label{tab:background_similarity}
\end{wraptable}
\subsubsection{Statistics.}As shown in Table \ref{tab:comp}, the \ourmethod dataset consists of $11,523$ video-question pairs from $2,000$ unique $15$-second videos. While longer videos are available, we limit the length to $15$ seconds to focus on manageable sequences and reduce challenges with longer content. Specifically, the dataset includes $4,000$ binary QA, $4,000$ multiple-choice questions (MCQ), and $3,523$ open-ended QA, all of which have been verified for quality.

Our adversarial pairs are categorized into non-scene-dependent cases (e.g., object/action edits with the same background) and scene-dependent cases (e.g., person as background). In non-scene-dependent cases, we compute full-frame metrics, while in scene-dependent cases, we mask the consistent subject and compute metrics on the masked regions. As shown in Table \ref{tab:background_similarity}, we use DINOv2, LPIPS, and SSIM metrics to assess background similarity. Evaluations are based on the keyframe at the start of each video clip, showing strong consistency across both types of cases and confirming high visual similarity across varying backgrounds.

\subsubsection{Benchmark Quality Assurance.} We adopt a compact, multi-stage pipeline that couples model-assisted curation with human review. Trained raters first vet GPT-4.1 story scripts for coherence and distinct, context-appropriate endings; annotators then verify each generated clip in Label Studio~\cite{LabelStudio} for description fidelity and baseline video quality. For adversarial pairs (cf. Fig.~\textcolor{magenta}{\ref{fig:video_level}}), reviewers retain only cases with \emph{highly similar backgrounds} and \emph{clear foreground semantic differences}, removing samples with background drift, foreground ambiguity, identity or attribute drift, temporal incoherence, or visible artifacts. Finally, for every validated pair we instantiate targeted QA templates spanning ten spatial/temporal hallucination types, and a separate rater group confirms strict question-video alignment for all $11,523$ QA instances. This four-level audit ensures that the generator never judges its own outputs and that retained errors can be traced to the intended foreground value or temporal-order change. Full protocols, rater instructions, and diagnostics are provided in the supplementary material.

\begin{figure}[t]
\centering
\includegraphics[width=\linewidth]{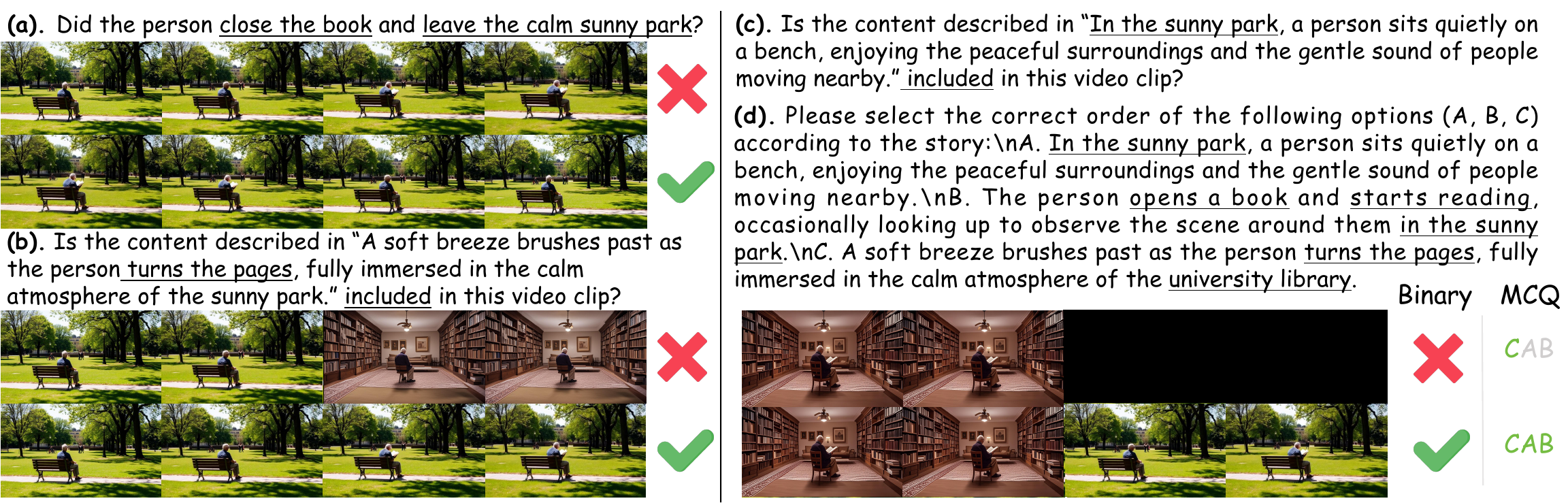}
\caption{Illustration of the increased challenge posed by temporal-level (\textbf{\textit{a}}) and spatial-level (\textbf{\textit{b}}) adversarial pairs compared to video-level adversarial pairs. Each spatial-level pair is accompanied by both binary QA (\textbf{\textit{c}}) and MCQ (\textbf{\textit{d}}) tasks. Notably, the MCQ format requires assessing and ranking the relevance of clip-level descriptions, thereby demanding a finer-grained understanding of spatial and temporal relationships within the video.}
\label{fig:clip_level}
\end{figure}

\subsection{Video-Level Adversarial Pairs} 
At the video level, we construct adversarial pairs by assembling positive and negative videos that differ in overall semantics or storyline as shown in Figure~\textcolor{magenta}{\ref{fig:video_level}}. The negative video in each pair is generated by replacing or altering the entire content, ensuring that its global meaning deviates from the positive counterpart. This design aims to test the model's ability to capture semantic consistency at a holistic level. Detecting such discrepancies is fundamental, as a reliable model should distinguish between videos with entirely different narratives, even when superficial visual similarities exist.

\subsection{Temporal-Level Adversarial Pairs} 
Temporal-level adversarial pairs are constructed by perturbing the temporal order of clips within a video. For negative samples, we mainly employ complete sequence reversal, since it preserves temporal semantics to the greatest extent and serves as the most fundamental way to test the model's capability in temporal reasoning and understanding. While other operations like shuffling or segment-level reordering can also disrupt chronological flow, sequence reversal is particularly suited for evaluating temporal logic as shown in Figure~\textcolor{magenta}{\ref{fig:clip_level}}(\textit{a}).

\subsection{Spatial-Level Adversarial Pairs} 

At the spatial level, we construct adversarial pairs by masking, replacing, or subtly modifying specific clips within a video, which introduces local inconsistencies such as missing objects or altered scenes. For example, changes to the book held by an elderly person or modifications in the surrounding environment, as shown in Figure~\textcolor{magenta}{\ref{fig:clip_level}}(\textit{b,c}), can disrupt object continuity while preserving the overall narrative structure. The binary QA is specifically designed to assess the model's sensitivity to spatial details and object-level coherence. Building on this, the MCQ format further evaluates the model's ability to maintain temporal consistency, addressing the common challenge where LVMs fail to accurately capture the foreground semantics of individual segments, resulting in contradictions or biases in temporal reasoning. By leveraging our proposed PairFlow framework, we can efficiently generate clip-level adversarial samples for fine-grained evaluation. Compared to clip-level captioning on real-world videos with fixed story logic, our method leverages customized story contexts to construct adversarial pairs with highly similar backgrounds and distinct foreground semantics. This design enables more targeted and lightweight evaluation of both spatial and temporal understanding.

\begingroup
\setlength{\tabcolsep}{1.6pt}
\renewcommand{\arraystretch}{0.98}

\begin{table*}[t]
\centering
\captionsetup{font=normalsize}
\caption{Performance comparison of open-source and closed-source LVMs on \ourmethod\ for binary, multi-choice, and open-ended QA. All metrics are reported as percentages. Human results serve as the upper bound. ``-'' denotes ``N/A''. \textbf{Bold} indicates best, \underline{underlined} second best.}
\label{tab:main}

{\scriptsize
\resizebox{0.95\textwidth}{!}{%
\begin{tabular}{lcccccccc}
\toprule
\multirow{2}{*}{\textbf{Method}} & \multirow{2}{*}{\textbf{Params}} & \multirow{2}{*}{\textbf{Language Model}}
& \multicolumn{3}{c}{\textbf{Binary}} 
& \multicolumn{2}{c}{\textbf{Multi-Choice}} 
& \textbf{Open-Ended} \\
\cmidrule(r){4-6} \cmidrule(r){7-8} \cmidrule(r){9-9}
& & & \textbf{wAcc $\uparrow$} & \textbf{FP ($\sim 0$)} & \textbf{Pct. Diff ($\sim 0$)}
& \textbf{F1 $\uparrow$} & \textbf{vAcc $\uparrow$ }
& \textbf{Desc. $\uparrow$} \\
\midrule
\rowcolor{lightgray!20} \multicolumn{9}{l}{{\textit{Open-source LVMs}}} \\
Video-ChatGPT \cite{maaz2023video} & 7B & LLaMA-7B        & 24.58 & 33.66 & 16.21 & 6.10 &  0.0 & 27.70 \\
Video-LLaVA \cite{lin2023video} & 7B & Vicuna-7B-v1.5   & 31.90 & 35.30 & 23.26 & 53.52 & 24.60 & 34.74 \\
VideoChat2 \cite{li2024mvbench} & 7B & Vicuna-7B-v0     & 18.47 &  \underline{2.12} & -45.89 & 38.03 & 0.79 & 42.25 \\
Video-LLaMA2 \cite{cheng2024videollama} & 7B & LLaMA2-7B     & 21.48 & 43.00 & 33.41 & 62.91 & 42.86 & 40.85 \\
Qwen2.5-VL-Instruct \cite{Qwen2.5-VL} & 7B & -              & \underline{41.66} & 48.71 & 14.77 & 61.50 & 42.86 & 45.07 \\
R1-OneVision \cite{yang2025r1onevision} & 7B & -              & 35.52 & 27.24 &  \underline{2.20} & 29.23 & 28.33 & 42.37 \\
ThinkLite-VL \cite{wang2025sota} & 7B & -                & 31.86 & 33.40 &  4.79 & 43.29 & 37.27 & 48.39 \\
ShareGPT4Video \cite{chen2024sharegpt4video} & 8B & LLaVA-Next-8B& 19.53 &  \textbf{1.67} & -44.95 & 38.50 & 0.08 & 34.27 \\
PLLaVA \cite{xu2024pllava} & 13B & Vicuna-13B-v1.5     & 32.06 & 45.72 & -6.74 & 8.92 &  2.38 & 41.31 \\
LLaMA-VID \cite{li2024llamavid} & 13B & Vicuna-13B-v1.5    & 18.79 & 53.13 & 49.75 & 33.80 & 15.08 & 30.52 \\
VILA1.5 \cite{lin2023vila} & 13B & -                     & 18.44 & 15.18 & -16.01 & 58.22 & 37.30 & 38.03 \\
\midrule
\rowcolor{lightgray!20} \multicolumn{9}{l}{{\textit{Closed-source LVMs}}} \\
Gemini-2.5-Flash \cite{team2024gemini}  & - & -         & 29.83 & 18.97 &  -8.82 & 62.44 & 39.68 & \underline{50.23} \\
Gemini-2.5-Pro \cite{team2024gemini}    & - & -         & \textbf{49.15} & 13.07 &  \textbf{-2.12} & \textbf{67.32} & \underline{43.36} & \textbf{54.68} \\
GPT-4o \cite{hurst2024gpt}              & - & -         & 26.97 & 29.16 & 10.97  & 59.15 & 38.10 & 47.89  \\
GPT-5-mini \cite{openai2025chatgpt5}    & - & -         & 29.33 & 19.65 &  3.28  & \underline{64.82} & \textbf{45.28} & 49.33  \\
\midrule
\rowcolor{yellow!20} Human    & - & -              & 74.32 & 9.28 & 4.37 & 89.21 & 79.66 & - \\
\bottomrule
\end{tabular}
}
}
\end{table*}

\endgroup

\section{Experiment}
\label{sec:experiment}
\subsection{Experimental Settings}

\paragraph{Models.} We evaluate $15$ mainstream LVMs spanning open-source systems and closed-source APIs. The model set covers widely used video-language baselines, recent reasoning-oriented open models, and frontier proprietary models. Table~\ref{tab:main} reports the full model list, citations, parameter sizes, and backbone details.

\paragraph{Evaluations.} For fair comparison, all models are evaluated on binary QA, MCQ, and open-ended description tasks.
For binary QA, to assess model robustness and performance, particularly in the context of adversarial video pairs and question pairs, we introduce several metrics: \textbf{Question Pair Accuracy}, \textbf{Video Pair Accuracy} and \textbf{Yes Percentage Difference (Pct. Diff)} as follows.

\textit{Question Pair Accuracy.}
This metric evaluates whether a model can consistently answer all instances within an adversarial pair correctly. Specifically, let $\mathbb{V}_i$ denote the set of videos in the $i$-th adversarial pair, and $\mathbb{Q}$ denote the set of question pairs. The Question Pair Accuracy is defined as:
\begin{equation}
  \mathrm{qAcc} = \frac{\sum_{i,k} \ind\left(\bigwedge_{V \in \mathbb{V}_i} b_{\mathcal{M}}(V, q(i,k))\right)}{|\mathbb{Q}|}, ~ b_{\mathcal{M}}(V, q) \in \{0, 1\}
\end{equation}
where $b_{\mathcal{M}}(V, q)$ is a binary indicator of correctness for model $\mathcal{M}$ on video $V$ and question $q$, and $\ind(\cdot)$ is the indicator function. This metric highlights the model's ability to provide consistent and robust answers across all elements of an adversarial pair, reflecting its resilience to both visual and textual adversarial perturbations.

\textit{Video Pair Accuracy.}
Symmetrically, let $\mathbb{V}\!=\!\bigcup_{i\in\mathcal{I}}\mathcal{V}_i$ be the set of videos across all adversarial pairs.
We say a specific video \(V\in\mathcal{V}_i\) is answered \emph{consistently} if the model is correct on \emph{all} questions in $\mathbb{Q}$ when evaluated on \(V\). The video-pair accuracy is defined as follows:
\begin{equation}
\mathrm{vAcc}
=\frac{1}{|\mathbb{V}|}\;
\sum_{i\in\mathcal{I}}\;\sum_{V\in\mathcal{V}_i}
\ind\!\left(\;\bigwedge_{q\in\mathbb{Q}} b_{\mathcal{M}}(V,q)\;\right).
\label{eq:vacc}
\end{equation}
To provide an intuitive summary of a model's overall robustness to video hallucination, we further report a weighted average, $\mathrm{wAcc}$, defined as:
\begin{equation}
    \mathrm{wAcc}=\frac{|\mathbb{Q}|\,\mathrm{qAcc}+|\mathbb{V}|\,\mathrm{vAcc}}{|\mathbb{Q}|+|\mathbb{V}|},
\end{equation}
where $|\mathbb{Q}|$ and $|\mathbb{V}|$ denote the numbers of question pairs and videos, respectively. This aggregates the two accuracies in proportion to their sample sizes.

\textit{Yes Percentage Difference.}
This metric measures the deviation between the proportion of ``yes'' responses given by the model and that in the ground truth. Formally,
\begin{equation}
 d_y = \frac{\sum_{(V, q) \in \mathcal{A}} \left[ \ind(\mathcal{M}(V, q) = \text{yes}) - \ind(y(V, q) = \text{yes}) \right]}{|\mathcal{A}|},
\end{equation}
where $\mathcal{A}$ is the set of all (video, question) pairs, $\mathcal{M}(V, q)$ is the model's answer, and $y(V, q)$ is the ground truth. A value of $|d_y|$ close to 1 indicates a strong bias towards a particular answer, while a value near 0 suggests balanced predictions.

For MCQ, besides \textit{qAcc}, we also report the F1 Score, a standard metric for multi-class classification:
\begin{equation}
 \text{F1} = 2 \times \frac{\text{Precision} \times \text{Recall}}{\text{Precision} + \text{Recall}},  
\end{equation}
where precision and recall are computed from true positives, false positives (FP), and false negatives.
\begin{figure*}[!t]
\centering
\includegraphics[width=\linewidth]{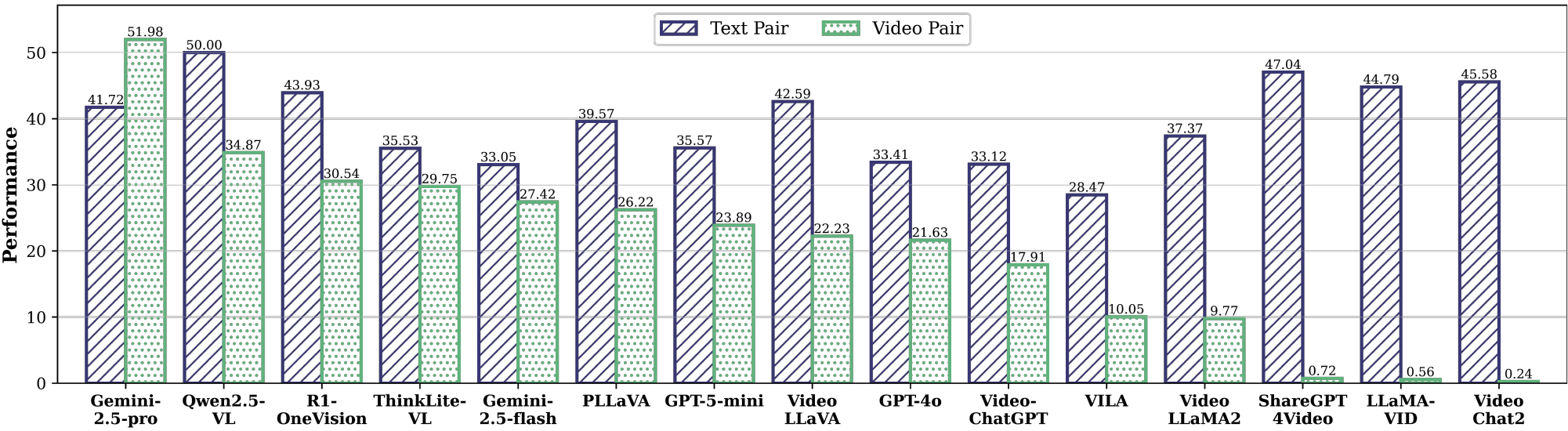}
\caption{Performance comparison between text-pair and video-pair hallucinations.}
\label{fig:twocol}
\end{figure*}

\begin{figure*}[ht]
  \centering
  \includegraphics[width=\linewidth]{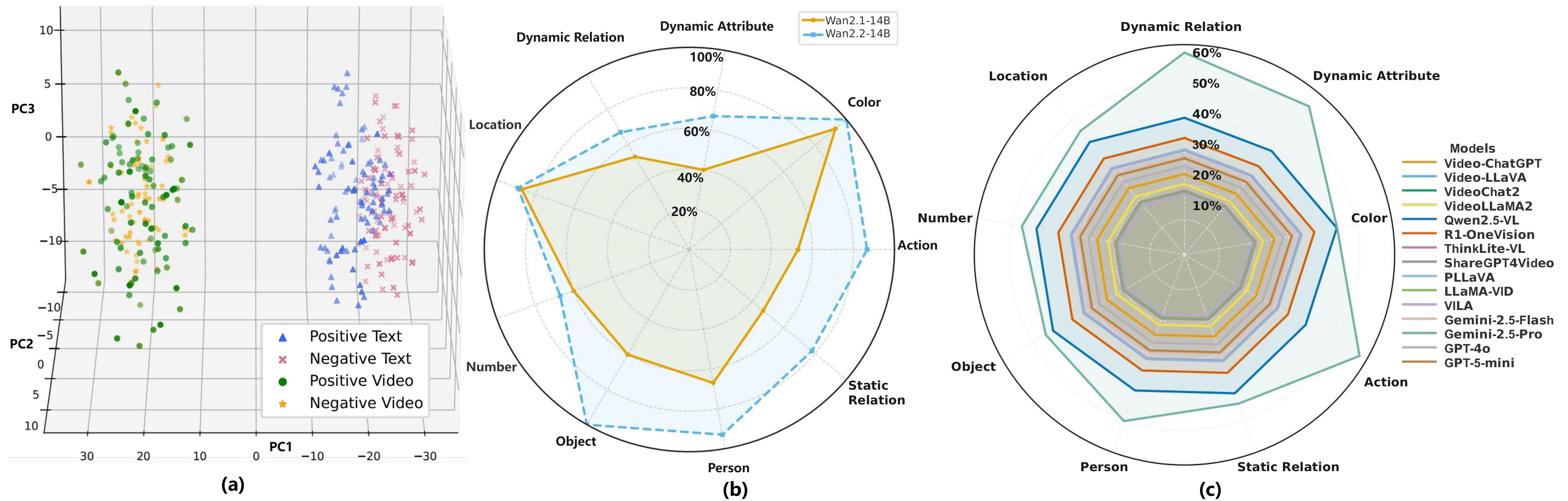}
  \caption{(\textbf{\textit{a}}) t-SNE of Qwen2.5-VL-Instruct shows overlap between positive and negative samples for video and text pairs, indicating the model barely separates relation polarity across modalities. (\textbf{\textit{b}}) Human verification rates for data synthesized by Wan 2.1 \textit{vs.} Wan 2.2, where Wan 2.2 achieves higher success. (\textbf{\textit{c}}) wACC on \ourmethod for 15 models, illustrating wide performance variability, and Gemini-2.5-Pro achieves the best place.}
  \label{fig:discussion}
\end{figure*}

\paragraph{Implementation Details.}
For story script generation, we utilized the GPT-4.1~\cite{hurst2024gpt} API to automatically produce narrative descriptions. Based on these scripts, we employed the advanced open-source text-to-image model FLUX~\cite{flux2024} to generate keyframes. Subsequently, precise foreground semantic editing was performed on these keyframes using the advanced image editing model SeedEdit~\cite{Shi2024SeedEditAI}. Leveraging both the edited images and the corresponding story segments, we further synthesized 480p video clips with the advanced open-source video generation model Wan2.1-14B and Wan2.2-14B~\cite{wan2025}. All data processing and inference were conducted on a cluster of eight H100 GPUs.
For open-ended questions evaluation, we employ GPT-4o~\cite{hurst2024gpt} and follow the settings in~\cite{zhang2024eventhallusion}. We recruited three English-proficient evaluators with computer science backgrounds to assess the benchmark results. To reduce bias, question-answer pairs were randomized to avoid consecutive similar types.

\subsection{Main Results}
\label{ssec:main_res}

As illustrated in Figure~\textcolor{magenta}{\ref{fig:twocol}}, video-pairs with highly similar backgrounds present a more significant challenge for models compared to text-pairs, often exacerbating hallucinations. As shown in Table~\ref{tab:main} and Figure~\textcolor{magenta}{\ref{fig:discussion}} (\textit{c}), closed-source LVMs outperform others, demonstrating superior calibration. Among these, Gemini-2.5-Pro stands out as the top performer, achieving the highest Binary wACC, the lowest FP, state-of-the-art Multi-choice results, and the most robust open-ended descriptions. GPT-5-mini, however, achieves the highest vAcc. The key advantage of low FP is its contribution to reducing spurious claims, an essential factor for safety-critical decision-making pipelines, where errors of commission can be costly. High wACC indicates broad resilience across adversarial video-text pairs and both clip- and video-level spatial semantics, rather than isolated successes. Among open-source models, Video-LLaMA2 excels in Multi-choice tasks (temporal reasoning), while Qwen2.5-VL achieves strong wACC, though with a higher FP, revealing a calibration gap. Overall, these results underscore the ongoing trade-off between spatial sensitivity and hallucination control. Bridging this gap will require synchronized advancements in temporal reasoning, spatial understanding, and risk-aware calibration.

\subsection{Further Analysis}
\label{ssec:discussion}

\paragraph{\textbf{Qualitative Results.}}
Figure~\textcolor{magenta}{\ref{fig:discussion}} (a) visualizes Qwen2.5 VL Instruct 7B embeddings for \ourmethod adversarial pairs. With strict background control, positive and negative samples largely overlap in both video and text, indicating weak separation of relation polarity. The learned representation underweights foreground semantics that are most informative for hallucination detection and instead appears biased toward background cues. This bias likely propagates to cross-modal alignment and blurs the distinction between genuine and adversarial pairs. In summary, the model captures scene context and style more than the subtle object and action evidence required for robust hallucination control.

\begin{figure*}[!t]
\centering
\includegraphics[width=\linewidth]{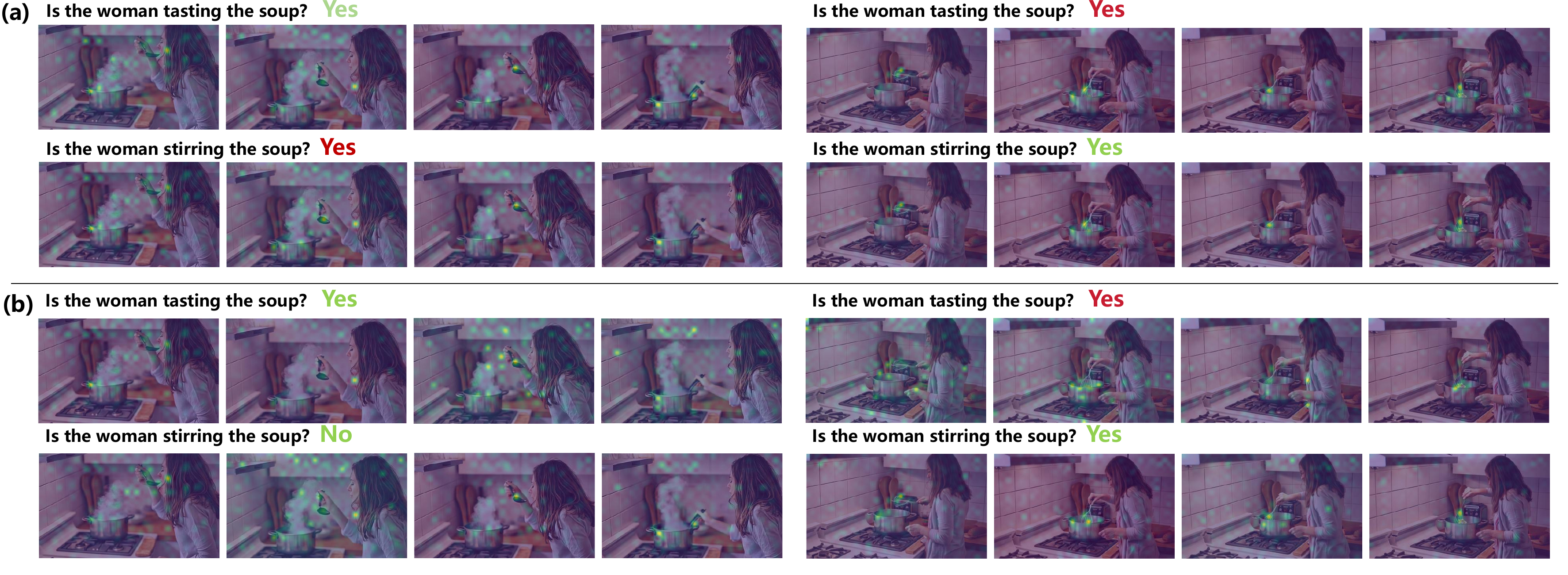}
\caption{Last-layer attention of (\textbf{\textit{a}}) Qwen2.5-VL-Instruct and (\textbf{\textit{b}}) ThinkLite-VL on adversarial text/video pairs with matched backgrounds and altered foreground actions. Heatmaps reveal focus patterns and alignment with correct \emph{vs.} incorrect predictions.}
\label{fig:cases}
\end{figure*}

\begin{figure*}[!t]
\centering
\includegraphics[width=0.94\linewidth]{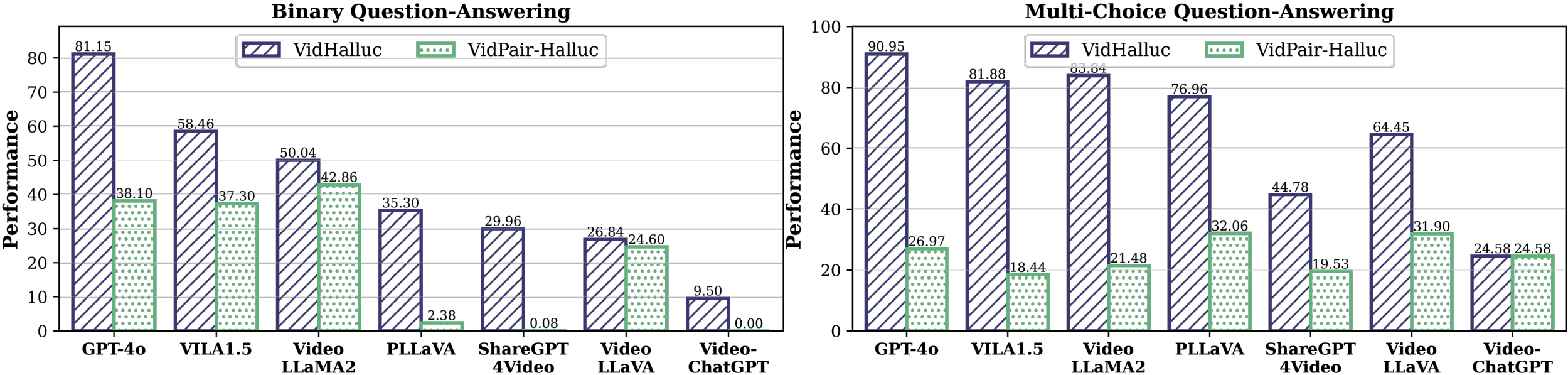}
\caption{Performance comparison of \textsc{VidHalluc} \textit{vs.} our \ourmethod.}
\label{fig:comp_with_other}
\end{figure*}
\begin{figure*}[!t]
    \centering
    \includegraphics[width=0.92\linewidth]{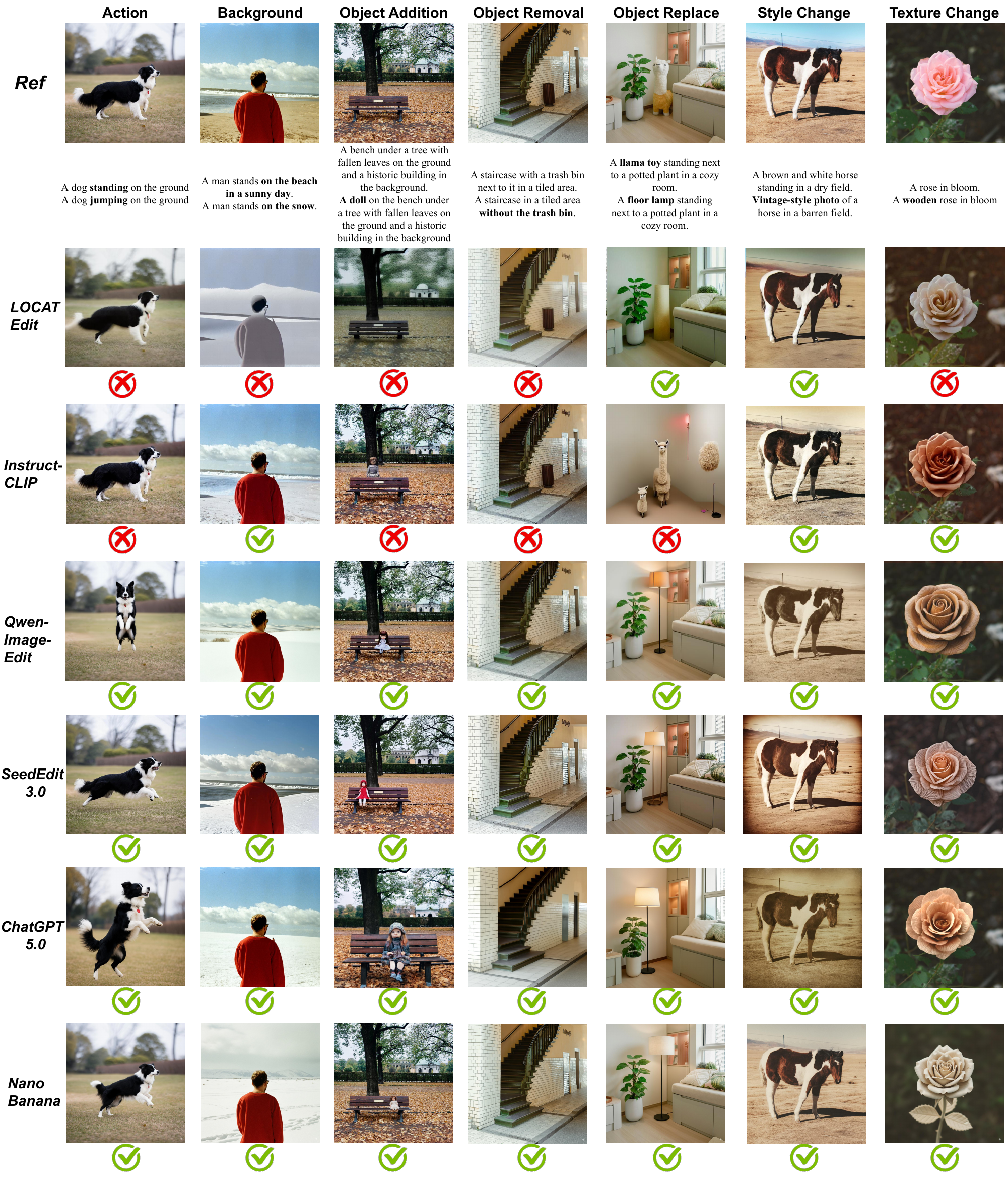}
  \caption{
Image editing performance comparison.
We compare LOCAT Edit~\cite{Soni2025LOCATEditGL}, Instruct-CLIP~\cite{Chen2025InstructCLIPII}, Qwen-Image-Edit~\cite{wu2025qwenimagetechnicalreport}, SeedEdit~3.0~\cite{Shi2024SeedEditAI}, ChatGPT~5.0~\cite{openai2025chatgpt5}, and Nano Banana~\cite{google2024nano_banana}; green/red marks indicate successful/failed edits. ChatGPT~5.0 and Nano Banana are the most consistent, followed by SeedEdit~3.0.
  }
    \label{fig:edit_comp}
\end{figure*}
\paragraph{\textbf{Synthesis Viability.}}
Figure~\textcolor{magenta}{\ref{fig:discussion}} (b) underscores the feasibility of generative data synthesis. As video generators advance, single-pass human vetting rates rise from roughly $63\%$ to about $83\%$, while Wan2.2 reaches approximately $70$--$100\%$ pass@1 across the evaluated axes. Complementary editing results in Figure~\textcolor{magenta}{\ref{fig:edit_comp}} suggest that stronger image and video generators will make \textsc{PairFlow} increasingly useful for both hallucination evaluation~\cite{bai2025impossible} and future contrastive or preference training~\cite{huang2025vistadpo,fu2025chip}.
\paragraph{\textbf{Case Study.}}
Following \cite{liu2025more}, we compare Qwen2.5-VL-Instruct with ThinkLite-VL by analyzing last-layer attention. Contrary to the claim in \cite{liu2025more}, when the task reduces to binary judgments without chain-of-thought requirements, the reasoning model consistently performs better. As shown in Figure~\textcolor{magenta}{\ref{fig:cases}} (\textit{b}), ThinkLite-VL exhibits richer, better-localized attention and more reliably separates subtle differences between adversarial video pairs, although both models remain vulnerable to adversarial text pairs.
\paragraph{\textbf{Performance Comparison with Related Benchmarks.}}
Unlike \textsc{VidHalluc}, \ourmethod constructs adversarial video pairs with highly similar backgrounds but significantly different foreground semantics. As shown in Figure~\textcolor{magenta}{\ref{fig:comp_with_other}}, most models suffer substantial degradation on \ourmethod compared to \textsc{VidHalluc}~\cite{li2024vidhalluc}, highlighting the difficulty of identifying fine-grained semantic differences when background cues are controlled.

\paragraph{\textbf{Scope and limitations.}}
Our benchmark complements real-video benchmarks: real videos capture natural diversity, while controlled pairs isolate queried foreground facts or temporal order for clearer attribution. The current release uses three $5$-second clips per video; longer compositions remain future work.
\section{Conclusion}
\label{sec:conclusion}
We introduce \ourmethod, a background-controlled benchmark for diagnosing video hallucination with adversarial pairs. \textsc{PairFlow} changes query-relevant foreground semantics while preserving backgrounds, enabling cleaner error attribution than mined videos. Results reveal persistent LVM weakness under fine-grained adversarial probes and motivate controlled pairs for diagnosis and alignment.

\clearpage
\appendix
\section{Additional Related Work}
\label{sec:app_related_work}
\subsection{Adversarial Video Pairs for Video Understanding}
Building upon powerful LLMs and integrating various multimodal encoders, recent research has led to the development of MLLMs and LVMs \cite{liu2024visual, fu2025vita, yin2024woodpecker, wu2024deepseek, li2023videochat, zhang2023video, lin2023video, li2024mvbench, cheng2024videollama, jin2024chat, li2025llama}. Through supervised fine-tuning (SFT) on visual instruction-tuning data, these models have achieved impressive multimodal understanding and significantly improved human-computer interaction. However, inheriting the intrinsic hallucination issues of LLMs, LVMs frequently produce hallucinations or fail to align their understanding of visual content with human intuition \cite{liu2024survey, zhang2024direct, li2024vidhalluc, liu2024seeing}. Increasing the scale of multimodal SFT data can alleviate these issues to some extent \cite{ahn2024tuning, tan2024evalalign, jiang2024supervised, yuan2024videorefer}, but this approach is often constrained by high annotation costs and computational demands, especially in video scenarios where data and training requirements are substantially greater. To address these challenges, the community has introduced preference alignment techniques such as DPO \cite{rafailov2024direct}, which align model outputs with human preference by leveraging pairs of preferred and rejected responses. Multimodal preference optimization extends this paradigm to visual and textual inputs, and has been widely adopted to enhance cross-modal alignment in MLLMs \cite{liu2024mia, xie2024v, zhou2024aligning,wu2025reverse,Wu2025SymmetricalVC,wang2024mdpo}. Recently, Hound-DPO \cite{zhang2024direct} successfully applied multimodal DPO to LVMs, improving video understanding and mitigating hallucinations, yet still overlooked the alignment of visual inputs. ~\cite{huang2025vistadpo,ding2025pami} further employ the visual preference pairs to enhance video-language alignment. Notably, VistaDPO \cite{huang2025vistadpo} introduces hierarchical preference optimization at the instance, temporal, and perceptive levels. The spatiotemporal adversarial video pairs within these preference pairs inherently contain subtle visual information, which enables VistaDPO to achieve significant improvement with relatively modest data volumes. 
However, the construction of such hierarchical preference pairs relies heavily on manual intervention to accurately annotate key frames and segments. This reliance, driven by the inherent complexity, fixed storylines, and spatiotemporal dependencies of real-world videos, significantly limits the scalability and efficiency of large-scale video preference data collection and makes annotation costly.
Due to the high cost of obtaining real-world video preference pairs, most existing approaches synthesize large numbers of QA pairs from a single video and construct misleading, hallucinated questions to generate textual preference data for training~\cite{zhang2024direct, huang2025vistadpo} and evaluating models' multimodal hallucination capabilities~\cite{wang2024videohallucer,zhang2024eventhallusion}. Thus, the effectiveness of LVMs trained and evaluated on video pairs remains underexplored.
Fortunately, advances in video generation technologies~\cite{kong2024hunyuanvideo,wan2025,magi1} combined with advanced editing techniques~\cite{zhuang2024task,zhu2024instantswap,Soni2025LOCATEditGL,Chen2025InstructCLIPII,Shi2024SeedEditAI} now enable precise control over spatial and temporal details, allowing the creation of fine-grained adversarial video pairs with plausible yet distinct variations.
These adversarial pairs have the potential to enable fine-grained video-language alignment and robust evaluation of multimodal reasoning~\cite{bai2025impossible}. Building on this potential, this work focuses on leveraging adversarial video pairs specifically for evaluation. We propose \textsc{PairFlow}, a novel data pipeline designed for high-quality adversarial video pairs, as well as \ourmethod for benchmarking LVMs for providing deeper insights into video hallucination.
\subsection{Video Hallucination Benchmarks}
LLMs gain strong language understanding from large-scale text pre-training, but their video encoders often lack similar representational strength. This mismatch can cause LLMs to generate confident yet incorrect outputs based on unreliable visual signals. To address this, researchers have introduced benchmarks that target various aspects of video hallucination for systematic diagnosis and mitigation.
Early benchmarks~\cite{Vript,zhang2024eventhallusion} mainly evaluate model understanding within single videos. These works use spatio-temporal questions to test basic comprehension of object relations, actions, and event sequences in isolated visual settings. For example, Vript-HAL~\cite{Vript} introduces long, richly annotated videos and designs tasks that target action and object hallucinations. EventHallusion~\cite{zhang2024eventhallusion} emphasizes event-level reasoning to reveal biases rooted in language priors.
Building on these foundations, later benchmarks introduce more adversarial textual contexts within single videos to test how robust models are against high-level semantic disturbances. VideoHallucer~\cite{wang2024videohallucer} follows this direction by systematically creating binary question pairs. Each pair includes \textit{``one factual, one hallucinated''} question, covering both intrinsic and extrinsic hallucination types. This approach allows for detailed analysis of how models handle misleading or counterfactual queries.
Recently, the evaluation paradigm has advanced further by using paired videos to better test a model's ability to align semantics and distinguish subtle visual differences. HallusionBench~\cite{guan2024hallusionbench} and \textsc{VidHalluc}~\cite{li2024vidhalluc} both create video pairs that are visually similar but semantically different. HallusionBench focuses on disentangling language hallucination from visual illusion, using carefully selected image-question pairs to separate knowledge priors from visual evidence. VidHalluc explicitly targets temporal hallucinations by assembling over 5,000 video pairs, evaluating models on their ability to distinguish actions, temporal sequences, and scene transitions. It uses CLIP~\cite{Radford2021LearningTV} and DINO~\cite{caron2021emerging} features to select semantically matched but visually diverse pairs, which directly exposes the limitations of current MLLMs in maintaining semantic consistency across related scenarios.

While these benchmarks have driven progress, the adversarial samples they use lack high visual similarity, leaving room for further improvement in sample quality. To further advance targeted, fine-grained hallucination evaluation, we enhance visual consistency across samples, enabling clearer assessment of a model's ability to capture subtle details in visual evidence.
\section{Limitations and Future Work}
\label{sec:app_limitations}
Despite the promising results, our proposed PairFlow pipeline is currently constrained by the limited physical priors embedded in existing video generation models, which poses a bottleneck for further large-scale scaling. Additionally, this work primarily explores scenarios with a small number of video clips. Future research is needed to investigate the generation and evaluation of longer compositional videos involving more segments, which would better reflect real-world complexity. Overall, our benchmark holds significant potential for extension, providing valuable insights for more fine-grained video-language alignment and broader applications in advancing large model capabilities for video understanding.
\section{Quality Assurance.}
\paragraph{Quality Control.}  
We employ trained participants to review all story scripts generated by GPT-4.1, ensuring each story is coherent and all candidate endings are distinct yet contextually appropriate. Using Label Studio~\cite{LabelStudio}, annotators then validate each generated video clip for consistency with its description and overall video quality. Further implementation and prompt details are provided below.
\paragraph{Human Validation of Adversarial Video Pairs.}  
We recruit a team of annotators to manually review selected adversarial video pairs, as illustrated by the qualitative examples in Figure~\textcolor{magenta}{\ref{fig:video_level}}. Annotators are instructed to identify and eliminate pairs with the following issues: (\textbf{\textit{i}}) insufficient or unclear foreground semantics in either video; (\textbf{\textit{ii}}) adversarial pairs that do not exhibit highly similar backgrounds and significant differences in foreground semantics. This process ensures that only high-quality adversarial pairs are retained.
\paragraph{QA Generation and Human Validation.}  
For each validated video pair, we generate a set of targeted QA instances spanning the 10 spatial and temporal hallucination categories defined in the benchmark, using category-specific question templates. A separate group of annotators reviews all generated QAs, ensuring each question is relevant and accurately matches the video content. This process guarantees a challenging and reliable QA set for model evaluation. 
\section{Prompt for Story Generation}
\subsection{Action Sample Generation}

\begin{tcolorbox}[breakable, colback=gray!10, colframe=black, title=Action Sample Generation, width=\linewidth]
\textbf{System Prompt}

You are an expert dataset designer for visual action understanding.  
Your task is to generate action event samples in JSON format.  
Each sample must meet the following strict requirements:

1. **Action Focus:**  
   Every sample describes a short, everyday scenario involving a human 
   or animal performing a specific, observable action.

2. **Segments:**  
   Each sample contains exactly three coherent segments (sentences), 
   describing the action's beginning, middle, and end.  
   The segments must be connected and form a natural mini-story.

3. **Candidate Values:**  
   For each sample, provide exactly three candidate values (verbs or 
   verb phrases) for the action, placed in the `{values}` slot in each segment.  
   - The three candidate values must correspond to **distinct, visually 
   distinguishable external actions** (e.g., "open", "close", "wash"), not just differences in purpose or mental state.
   - Substituting any candidate value into the segments must result in a 
   natural, logical, and visually clear story.
   - Avoid subtle or abstract distinctions (e.g., "compose"/"play"
   /"record"); prefer actions with clear physical differences.

4. **Formatting:**  
   Output a JSON array called `"action"`, where each element contains:
   - `"id"`: a unique identifier (e.g., "action\_0001")
   
   - `"segments"`: an array of three English sentences, each containing a `{values}` placeholder
   
   - `"values"`: an array of three candidate action values (verbs or verb phrases)

5. **Diversity:**  
   Ensure the actions, scenarios, and candidate values cover a wide range of everyday activities and are not repetitive.

6. **Language:**  
   All content must be in fluent, idiomatic English.

7. **Sample Size:**  
   Generate exactly 40 distinct samples per request.

Return only the JSON content as specified, with no additional commentary.

\vspace{1.5mm}
\textbf{User Prompt}

Please generate 40 action event samples according to the following requirements:

- Each sample should describe a short, everyday scenario in three connected English segments, with a `{values}` placeholder for the action.
- For each sample, provide three candidate action values (verbs or verb phrases) that are visually and physically distinct from each other, so that substituting any value results in a clear, observable difference in the described action.
- Ensure the segments remain natural and logical with any candidate value substituted.
- Output the results as a JSON array named `"action"`, with each element containing `"id"`, `"segments"`, and `"values"` as described.
- All content must be in English.
- Generate exactly 40 samples.

Return only the JSON content.
\end{tcolorbox}
\subsection{Dynamic Visual Attribute Sample Generation}

\begin{tcolorbox}[breakable, colback=gray!10, colframe=black, title=Dynamic Visual Attribute Sample Generation, width=\linewidth]
\textbf{System Prompt}

You are a high-quality visual dynamic attribute sample generator.  
Your goal is to batch-generate samples for visual dynamic attribute scenarios according to the user's instructions.  
Each sample should include the following elements:

1. **ID**: A unique identifier (e.g., dynamic\_attribute\_0001).

2. **segments**: 3 descriptive sentences forming a coherent, intuitive visual story that depicts the dynamic change of an 
attribute, suitable for illustration via image/video.  

3. **values**: 3 candidate values with significant distinction, perfectly fitting the story context. All values must represent clearly visible attribute changes (such as speed, quantity, distance, angle, color, shape, size, brightness, occlusion, density, etc.) and should avoid abstract or hard-to-visualize descriptions.

The generated samples must satisfy:

- Attribute change trends are clear and easy to visualize.

- Candidate values are highly distinguishable, with no ambiguous or hard-to-differentiate terms.

- Content is diverse and non-repetitive.

- Use concise, accurate English descriptions.

- Output must be in standard JSON structure, with the key named "dynamic\_attribute" and the value being an array of samples.

Do not output anything except the JSON content.

\vspace{1.5mm}
\textbf{User Prompt}

Please batch-generate N high-quality samples (where N is specified by the user, e.g., 100) for a visual dynamic attribute recognition task. Each sample must include:
- id: A unique identifier (e.g., dynamic\_attribute\_0001).
- segments: 3 sentences forming a coherent, intuitive visual story describing the dynamic change of one attribute.
- values: 3 candidate values that are highly distinguishable and clearly reflected in the scenario, tightly coupled with the story context.

The attribute types should cover speed, quantity, distance, angle, color, shape, size, brightness, occlusion, density, and other intuitive visual properties. Avoid abstract or hard-to-visualize descriptions.

The required output format is as follows:
\begin{lstlisting}[basicstyle=\ttfamily\footnotesize
, breaklines=true, frame=single]
{
  "dynamic_attribute": [
    {
      "id": "dynamic_attribute_0001",
      "segments": [
        "A runner starts on a city track, his speed {values} as he passes under streetlights.",
        "Midway through the race, his running pace {values}, sweat visible on his brow.",
        "At the finish line, his speed {values}, and he raises his arms in triumph."
      ],
      "values": [
        "gradually increases",
        "gradually decreases",
        "remains steady"
      ]
    },
    ...
  ]
}
\end{lstlisting}

Please output only the JSON content, and do not include any text other than the JSON.
\end{tcolorbox}
\subsection{Sequence Sample Generation}
\begin{tcolorbox}[breakable, colback=gray!10, colframe=black, title=Sequence Sample Generation, width=\linewidth]
\textbf{System Prompt}

You are an expert data generator for visual storytelling datasets.  
Your task is to generate story sequence data for image or video generation.  
Each story consists of three segments describing a simple, everyday event or activity.  
For each story, provide three different orderings of the same three detailed actions, where each action includes vivid scene details suitable for visual generation.  
Return the result in JSON format as shown in the example.

\vspace{1.5mm}
\textbf{User Prompt}

Generate N story event samples as described above (replace N with the required number, e.g., 40).  
Each sample should have:
- An "id" field (e.g., "sequence\_0001").

- A "segments" array of three English sentences, each with a placeholder {values[0]}, {values[1]}, {values[2]} for the actions.

- A "values" field: a 2D array with three permutations of the same three actions, each action being a richly detailed, visually descriptive phrase.

- All actions should fit naturally into the segment sentences and be suitable for generating realistic images or videos.

\vspace{1.5mm}
\textbf{Example Format:}

\begin{lstlisting}[basicstyle=\ttfamily\footnotesize
, breaklines=true, frame=single]
{
  "sequence": [
    {
      "id": "sequence
_
0001",
      "segments": [
        "The child first {values[0]}.",
        "Then, the child {values[1]}.",
        "Finally, the child {values[2]}."
      ],
      "values": [
        [
          "walks slowly across the grassy field, looking around curiously",
          "runs quickly past the playground, his arms swinging with excitement",
          "jumps high over a small puddle, landing with a bright smile"
        ],
        [
          "jumps high over a small puddle, landing with a bright smile",
          "walks slowly across the grassy field, looking around curiously",
          "runs quickly past the playground, his arms swinging with excitement"
        ],
        [
          "runs quickly past the playground, his arms swinging with excitement",
          "jumps high over a small puddle, landing with a bright smile",
          "walks slowly across the grassy field, looking around curiously"
        ]
      ]
    }
    // ...more samples
  ]
}
\end{lstlisting}

\vspace{1.5mm}
\textbf{Requirements:}
- Each story's three actions must be unique, richly detailed, and visually specific.
- Each of the three "values" arrays must be a different permutation of the same three actions.
- The language should be vivid and descriptive to support high-quality image/video generation.
- Do not repeat actions or use generic verbs without scene context.

Generate N such samples in the specified format.
\end{tcolorbox}
\section{Prompt for QA Generation}
\subsection{Multiple-Choice Question}

\begin{tcolorbox}[breakable, colback=gray!10, colframe=black, title=Multiple-Choice Question Generation, width=\linewidth]
\textbf{System Prompt}

You are an expert at generating multiple-choice question and answer (MCQ) data for multimodal datasets. Given a video segment description, a question, and a set of answer choices, your task is to select the most appropriate answer based on the context. Always ensure your answer is accurate and based on the information provided in the segment description.

\vspace{1.5mm}
\textbf{User Prompt}

Given the following video segment description, question, and answer choices, select the single best answer. Respond only with the letter corresponding to the correct choice (e.g., "A").

\vspace{1.5mm}
\textbf{Video Segment Description:}
\{used\_segment\}

\textbf{Question:}
\{question\}

\textbf{Choices:}
A. \{choice\_A\}
B. \{choice\_B\}
C. \{choice\_C\}
D. \{choice\_D\}

\vspace{1.5mm}
\textbf{Instructions}

\begin{enumerate}
    \item Read the segment description carefully.
    \item Choose the answer that best fits the context.
    \item Only respond with the letter of the correct answer (e.g., \textbf{A}).
\end{enumerate}

\vspace{1.5mm}
\textbf{Example}

\textbf{Video Segment Description:}
The child picks up a balloon, ready to inflate it.

\textbf{Question:}
What action was the child getting ready to do with the balloon?

\textbf{Choices:}
A. inflate  
B. pop  
C. paint  
D. tie  

\textbf{Answer:}  
A
\end{tcolorbox}
\subsection{Binary Question}

\begin{tcolorbox}[breakable, colback=gray!10, colframe=black, title=Binary Question Generation, width=\linewidth]
\textbf{System Prompt}

You are an expert at generating binary (Yes/No) question-answer data for multimodal datasets. Given a video segment description and a related Yes/No question, your task is to determine the correct answer based on the information provided. Your answer should be either "Yes" or "No", strictly according to the context.

\vspace{1.5mm}
\textbf{User Prompt}

Given the following video segment description and Yes/No question, answer with either "Yes" or "No" based on the context.

\vspace{1.5mm}
\textbf{Video Segment Description:}
\{used\_segment\}

\textbf{Question:}
\{question\}

\vspace{1.5mm}
\textbf{Instructions}

\begin{enumerate}
    \item Carefully read the segment description.
    \item Answer the question strictly based on the information provided.
    \item Respond only with "Yes" or "No".
\end{enumerate}

\vspace{1.5mm}
\textbf{Example}

\textbf{Video Segment Description:}
The child picks up a balloon, ready to inflate it.

\textbf{Question:}
Did the child prepare to blow air into the balloon?

\textbf{Answer:}
Yes
\end{tcolorbox}
\subsection{Open-Ended Question}

\begin{tcolorbox}[breakable, colback=gray!10, colframe=black, title=Open-Ended Question Generation, width=\linewidth]
\textbf{System Prompt}

You are an expert at generating open-ended question-answer (QA) pairs for multimodal datasets. Given a segment description from a video and a related question, your task is to generate a natural, contextually appropriate, and informative answer in English. The answer should accurately reflect the action or event described, and provide concise reasoning or context when appropriate. Use natural and fluent English, and avoid repetition.

\vspace{1.5mm}
\textbf{User Prompt}

Given the following video segment description and question, generate a natural, open-ended answer in English:

\vspace{1.5mm}
\textbf{Video Segment Description:}
\{used\_segment\}

\textbf{Question:}
\{question\}

\vspace{1.5mm}
\textbf{Instructions:}
\begin{enumerate}
    \item Your answer should be contextually relevant and reflect the action or event described in the segment.
    \item Use natural, fluent English, and avoid repeating the question verbatim.
    \item Provide a concise explanation or reasoning behind the action if possible.
    \item Use the correct verb tense and pronouns based on the segment.
    \item The answer should be a self-contained, informative sentence or short paragraph.
\end{enumerate}

\vspace{1.5mm}
\textbf{Example}

\textbf{Video Segment Description:}
The child picks up a balloon, ready to inflate it. She holds the balloon and inflates it with excitement. Afterward, she smiles, having just inflated the balloon.

\textbf{Question:}
What action was the child getting ready to do with the balloon?

\textbf{Answer:}
The child was preparing to inflate the balloon, clearly excited about the activity. After inflating it, she smiled, showing her enjoyment of the moment.
\end{tcolorbox}

\subsection{Supplementary Frontier Results}
\label{app:frontier_models}

To complement the main-paper comparison in Table~\ref{tab:main}, we further evaluate several newly released flagship frontier models in the appendix, including Gemini-3.1-Pro~\cite{deepmind2026gemini}, GLM-5~\cite{zhipu2025glm5}, GPT-5.4~\cite{openai2026gpt}, Kimi-2.5~\cite{moonshot2026kimi}, MiniMax-M2.5~\cite{minimax2026m2}, Opus-4.6~\cite{anthropic2026opus}, and Qwen3.5-397B-A17B~\cite{qwen2026qwen35}. This setup preserves the historical alignment of the main text while revealing a clear forward trend: recent frontier LVMs have moved substantially closer to human reference performance on our benchmark, yet still exhibit distinct trade-offs in hallucination sensitivity, fine-grained visual grounding, and descriptive ability.

\begingroup
\setlength{\tabcolsep}{2.2pt}
\renewcommand{\arraystretch}{0.98}
\begin{table*}[!t]
\centering
\captionsetup{font=normalsize}
\caption{Supplementary comparison on newly released flagship frontier models. In contrast to the main-paper benchmark in Table~\ref{tab:main}, these models form a much stronger frontier tier and collectively narrow the gap to human reference performance, while still showing complementary strengths rather than a single uniformly dominant behavior across binary, multi-choice, and open-ended settings. All metrics are reported as percentages. ``-'' denotes ``N/A''. \textbf{Bold} indicates best, \underline{underlined} second best.}
\label{tab:frontier_appendix}
{\scriptsize
\resizebox{0.78\textwidth}{!}{%
\begin{tabular}{lcccccc}
\toprule
\multirow{2}{*}{\textbf{Method}}
& \multicolumn{3}{c}{\textbf{Binary}}
& \multicolumn{2}{c}{\textbf{Multi-Choice}}
& \textbf{Open-Ended} \\
\cmidrule(r){2-4} \cmidrule(r){5-6} \cmidrule(r){7-7}
& \textbf{wAcc $\uparrow$} & \textbf{FP ($\sim 0$)} & \textbf{Pct.\ Diff ($\sim 0$)}
& \textbf{F1 $\uparrow$} & \textbf{vAcc $\uparrow$} & \textbf{Desc.\ $\uparrow$} \\
\midrule
Gemini-3.1-Pro \cite{deepmind2026gemini} & \textbf{63.84} & \underline{10.12} & \textbf{-0.86} & \underline{78.94} & 58.63 & \textbf{67.91} \\
GLM-5 \cite{zhipu2025glm5} & 51.72 & 17.84 & 3.47 & 72.16 & 52.74 & 58.46 \\
GPT-5.4 \cite{openai2026gpt} & \underline{62.91} & 11.36 & \underline{0.94} & \textbf{80.52} & \textbf{61.24} & \underline{66.73} \\
Kimi-2.5 \cite{moonshot2026kimi} & 55.38 & 15.92 & 2.41 & 76.44 & 54.87 & 62.95 \\
MiniMax-M2.5 \cite{minimax2026m2} & 49.86 & 19.73 & 5.28 & 71.38 & 50.46 & 57.18 \\
Opus-4.6 \cite{anthropic2026opus} & 56.31 & \textbf{9.61} & 2.04 & 77.12 & 56.82 & 64.32 \\
Qwen3.5-397B-A17B \cite{qwen2026qwen35} & 58.42 & 11.74 & 1.72 & 75.93 & \underline{59.48} & 60.85 \\
\midrule
Human & 74.32 & 9.28 & 4.37 & 89.21 & 79.66 & - \\
\bottomrule
\end{tabular}}
}
\end{table*}
\endgroup

\begin{figure*}[!t]
\centering
\includegraphics[width=\linewidth]{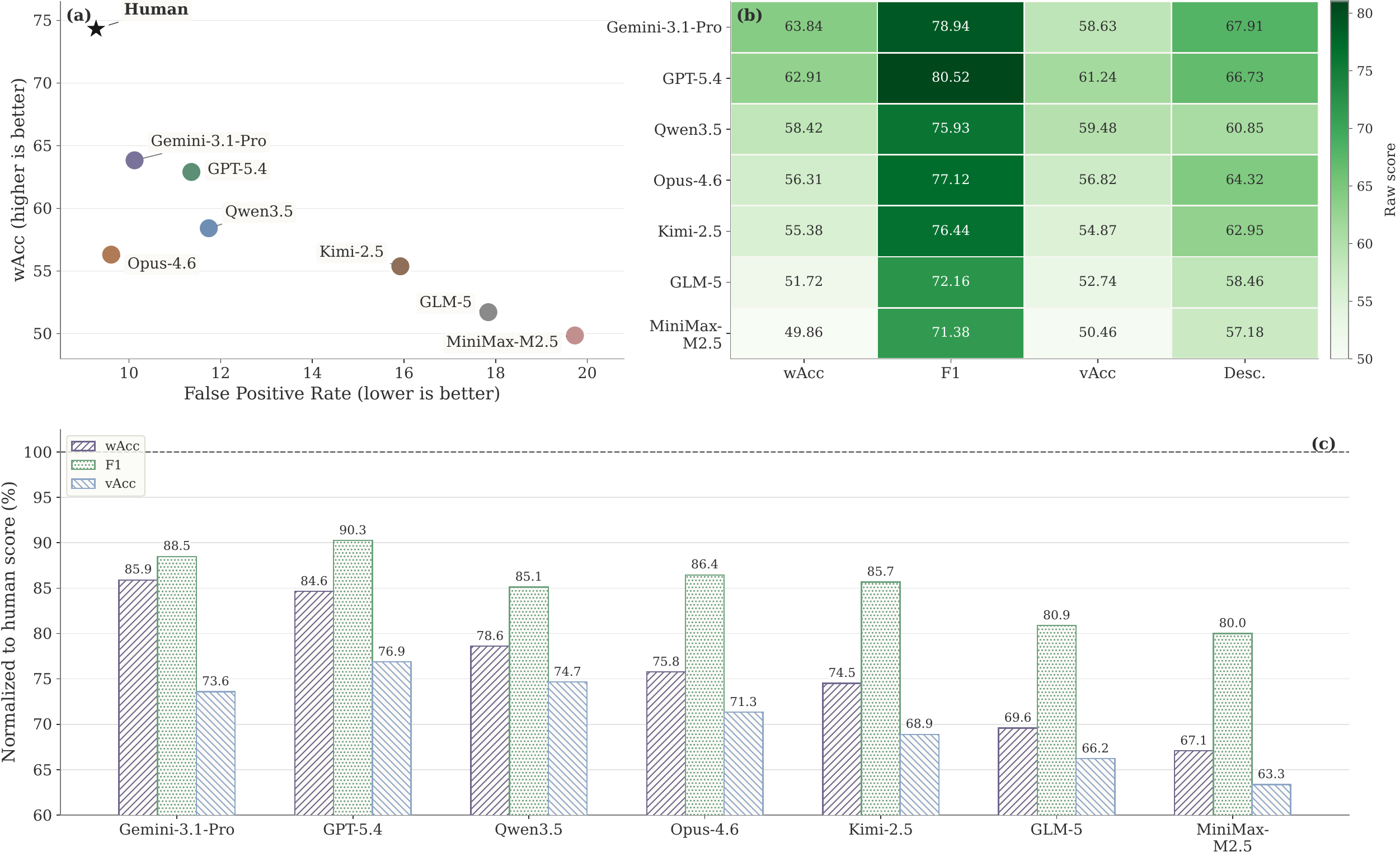}
\caption{Supplementary insights for newly released flagship closed-source models. (\textbf{a}) All seven frontier models occupy different positions in the robustness--conservativeness plane, showing that stronger binary robustness (wAcc) is achieved through different trade-offs with false-positive control (FP). (\textbf{b}) A score heatmap across all seven models further shows that the frontier is complementary rather than uniformly dominant: Gemini-3.1-Pro leads binary robustness and description, GPT-5.4 leads F1 and vAcc, while the remaining models retain competitive profiles on selected dimensions. (\textbf{c}) After normalizing each metric by its corresponding human score, all seven models can be compared on a common reference scale, showing that the strongest frontier systems already recover a large fraction of human reference performance on wAcc, F1, and vAcc.}
\label{fig:frontier_appendix}
\end{figure*}

The appendix results echo the main-text conclusion that closed-source models are generally better calibrated, but they also show that the frontier has advanced noticeably beyond the 15-model benchmark reported in the main paper. In particular, Gemini-3.1-Pro and GPT-5.4 emerge as a tightly matched leading pair: Gemini-3.1-Pro attains the best binary robustness, with the highest wAcc and the smallest Pct.\ Diff, while GPT-5.4 achieves the strongest multi-choice performance, obtaining the best F1 and vAcc. This pattern suggests that Gemini-3.1-Pro is slightly better calibrated for global hallucination discrimination and open-ended description, whereas GPT-5.4 is stronger at fine-grained visual grounding and answer selection. Notably, Opus-4.6 yields the lowest FP among the frontier closed-source models, indicating a more conservative prediction profile that reduces over-triggered hallucination judgments. Qwen3.5 also reaches the second-best vAcc, and Kimi-2.5 remains competitive on F1, reinforcing that recent flagship models are converging toward human-level behavior through different capability profiles rather than a single common operating point.

Figure~\ref{fig:frontier_appendix} provides a more structured view of the emerging frontier pattern using all seven flagship models. In Figure~\ref{fig:frontier_appendix}(\textbf{a}), the trade-off plot shows that better hallucination robustness is not achieved by a single strategy: Opus-4.6 stays closest to the low-FP regime, whereas Gemini-3.1-Pro and GPT-5.4 accept slightly higher false-positive rates in exchange for substantially stronger overall robustness. In Figure~\ref{fig:frontier_appendix}(\textbf{b}), the heatmap makes the complementary nature of the frontier especially clear, showing that Gemini-3.1-Pro leads binary robustness and description while GPT-5.4 leads F1 and vAcc, with the remaining models retaining competitive but distinct profiles. In Figure~\ref{fig:frontier_appendix}(\textbf{c}), the normalized human-reference comparison shows that, after dividing each score by the corresponding human score for that metric, the strongest frontier systems recover a large fraction of human reference performance on wAcc, F1, and vAcc, while the rest still form a relatively compact and competitive tier. Taken together, these three views strengthen the appendix conclusion that hallucination-sensitive video understanding now differentiates top-tier models by their capability profiles and calibration styles, rather than by a simple strong-versus-weak separation.

\clearpage
\IfFileExists{main_arxiv.bbl}{%

}{%
\bibliographystyle{splncs04}
\bibliography{main}
}

\end{document}